\documentclass[letterpaper,journal]{IEEEtran}

\usepackage[utf8]{inputenc} 
\usepackage[T1]{fontenc}    
\usepackage{url}            
\usepackage{cite}           
\usepackage{booktabs}       
\usepackage{amsfonts}       
\usepackage{nicefrac}       
\usepackage{microtype}      
\usepackage{xcolor}         

\usepackage{multirow}
\usepackage{colortbl}

\usepackage{amsthm,amsmath,amssymb}
\usepackage{mathrsfs}
\usepackage{graphicx}
\usepackage{booktabs}
\usepackage{capt-of}
\usepackage{algorithm}
\usepackage{algpseudocode}
\usepackage{hyperref}       
\hypersetup{hidelinks}

\usepackage{wrapfig}
\usepackage{caption}

\definecolor{eccvblue}{rgb}{0.12,0.49,0.85}
\definecolor{cvprblue}{rgb}{0.21,0.49,0.74}
\definecolor{color3}{gray}{0.95}
\definecolor{rouse}{rgb}{0.981,0.961,0.941}

\title{SlowFast-SCI: Slow-Fast Deep Unfolding Learning for Spectral
Compressive Imaging}

\author{%
  Haijin Zeng$^{1,*}$\thanks{$^*$Equal contribution.
  \{\href{mailto:haijin.zeng2018@gmail.com}{haijin.zeng2018, }
  \href{mailto:xuanlu1113@gmail.com}{xuanlu1113}\}\href{@gmail.com}{@gmail.com}
  },
  Xuan Lu$^{1,6,*,\dagger}$,
  Jiezhang Cao$^{2}$,
  Kai Zhang$^{3}$,
  Yurong Zhang$^{2}$,
  Qiangqiang Shen$^{4}$,
  Guoqing Chao$^{5}$,
  Li Jiang$^{6}$, 
  Yongyong Chen$^{1,\dagger}$,
  Jingyong Su$^{1}$, and
  Jie Liu$^{1}$
  \thanks{$^\dagger$Corresponding author. \href{mailto:xuanlu1113@gmail.com}{xuanlu1113@gmail.com}, \href{mailto:cyy2020@hit.edu.cn}{cyy2020@hit.edu.cn}}%
  \thanks{$^{1}$Harbin Institute of Technology, Shenzhen;
  $^{2}$Shanghai Jiaotong University; $^{3}$Nanjing University; $^{4}$City University of Hong Kong;
  $^{5}$Harbin Institute of Technology, Weihai; $^{6}$The Chinese University of Hong Kong, Shenzhen.}
}

\begin{document}

\maketitle

\begin{abstract}
Humans learn in two complementary ways: a slow, cumulative process that builds broad, general knowledge, and a fast, on-the-fly process that captures specific experiences. Existing deep-unfolding methods for spectral compressive imaging (SCI) mirror only the slow component—relying on heavy pre-training with many unfolding stages—yet they lack the rapid adaptation needed to handle new optical configurations. As a result, they falter on out-of-distribution cameras, especially in bespoke spectral setups unseen during training. This depth also incurs heavy computation and slow inference.
To bridge this gap, we introduce SlowFast-SCI, a dual-speed framework seamlessly integrated into any deep unfolding network beyond SCI systems. During slow learning, we pre-train or reuse a priors-based backbone and distill it via imaging guidance into a compact fast-unfolding model. In the fast learning stage, lightweight adaptation modules are embedded within each stage and fine-turned self-supervised at test time via a self-supervised loss—without retraining the backbone. 
To the best of our knowledge, SlowFast-SCI is the first test‐time adaptation–driven deep unfolding framework for efficient, self‐adaptive spectral reconstruction. Its dual‐stage design unites offline robustness with on‐the‐fly per‐sample calibration—yielding over 70\% reduction in parameters and FLOPs, up to 5.79 dB PSNR improvement on out‐of‐distribution data, preserved cross‐domain adaptability, and a 4× faster adaptation speed. 
In addition, its modularity integrates with any deep-unfolding network, paving the way for self-adaptive, field-deployable imaging and expanded computational imaging modalities. 
\emph{Code is available in Supplementary Material.} 
\end{abstract}

\section{Introduction} \label{sec:intro}

Humans are endowed with a complementary learning system \cite{mcclelland1995there, hong2024slowfast} that operates on two distinct timescales. A slow process in the neocortex gradually integrates diverse experiences into a coherent world model, enabling robust prediction of action outcomes \cite{schwesinger55review}. 
In parallel, a fast process in the hippocampus rapidly encodes episodic details, supporting agile adaptation to novel circumstances \cite{tulving1983elements}. 
Integrating such a dual-speed mechanism into neural networks enables rapid adaptation, \textit{i.e.}, swift recalibration to new domains or unseen inputs—thereby enhancing both flexibility and generalization.


This dual-speed learning paradigm inspires advances in spectral compressive imaging (SCI), a computational photography technique that aims to reconstruct rich hyperspectral information from compressed measurements. 
The predominant SCI architecture, coded aperture snapshot spectral imaging (CASSI) \cite{gehm2007single_cassi, wagadarikar2008single_cassi_sparsity, meng2020endtsa_cassi}, employs a spatially coded mask and a dispersive element to modulate, shift, and integrates spectral bands onto a two-dimensional sensor \cite{zhang2022herosnet}, achieving high-throughput, cost-effective, and bandwidth-efficient acquisition. 
The resulting hyperspectral images (HSIs) deliver rich spatial–spectral signatures that significantly enhance scene characterization beyond conventional RGB imagery. Consequently, SCI has found broad applications across computer vision and remote sensing, \textit{e.g.}, object detection \cite{van2010tracking_object, kim20123d_object}, medical imaging \cite{lu2014medical, Meng_2020_OL_SMEM_medical}, and land-cover classification \cite{ borengasser2007hyperspectral_remote, yuan2017hyperspectral_remote}.

\begin{figure}[t]
    \centering
    \includegraphics[width=1\linewidth]{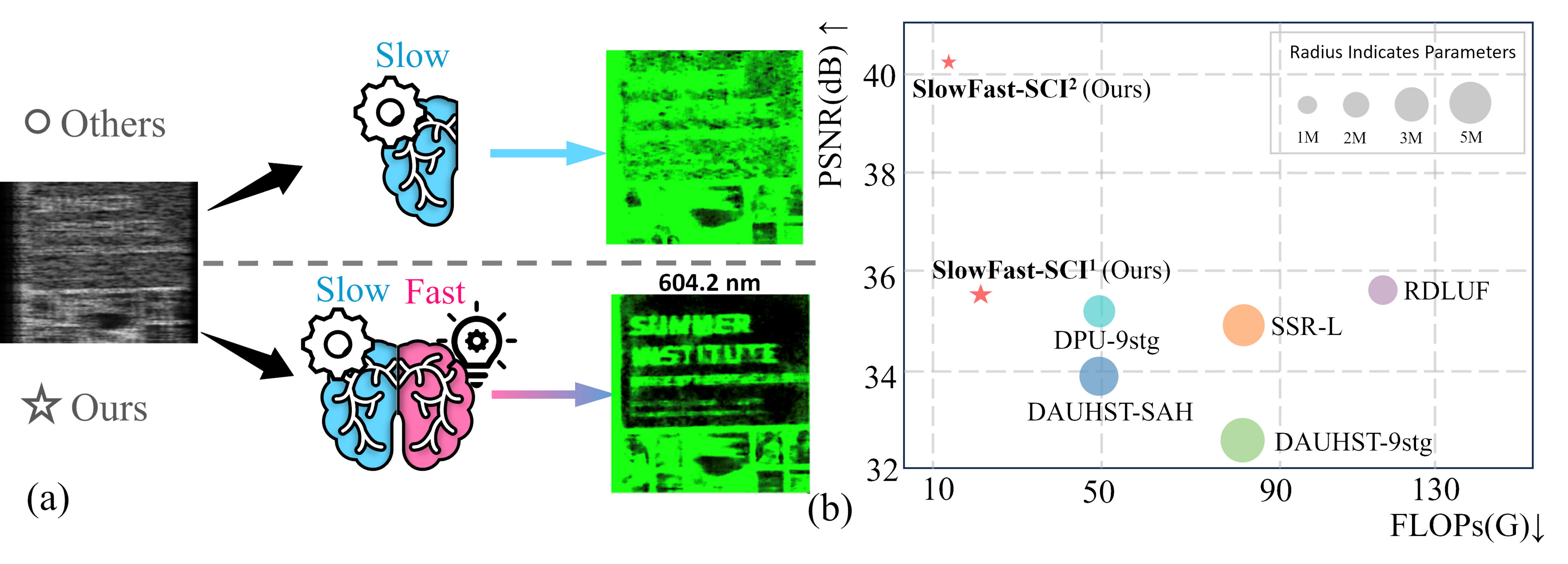}
    \vspace{-6mm}
    \caption{(a) Previous SCI reconstruction method vs. our SlowFast-SCI. (b) The PSNR-FLOPs-Params analysis comparing the proposed SlowFast-SCI with latest state-of-the-art methods.}
    \label{fig:placeholder}
    \vspace{-4mm}
\end{figure}

Based on the CASSI acquisition model, a broad spectrum of reconstruction algorithms has been proposed to recover a three-dimensional hyperspectral data cube from its two-dimensional measurement. Traditional model-based methods employ handcrafted priors—such as sparsity \cite{wagadarikar2008single_cassi_sparsity,kittle2010multiframe_sparsity}, total variation \cite{rudin1992nonlinear_tv,wang2015dual_tv}, and low-rank constraints \cite{liu2018rank_lowrank,zhang2019computational_lowrank}—but they require careful parameter tuning and often yield suboptimal quality and slow convergence. The advent of deep learning introduced end-to-end networks that significantly improve both reconstruction accuracy and speed \cite{mst,cst,dauhst,hdnet,meng2020gap,meng2020endtsa_cassi,bisci,Dong_2023_CVPR_rdluf,Zhang_2024_CVPR_ssr,Zhang_2024_CVPR_dpu}.

\begin{figure*}[t]
    \centering
    \includegraphics[width=1\linewidth]{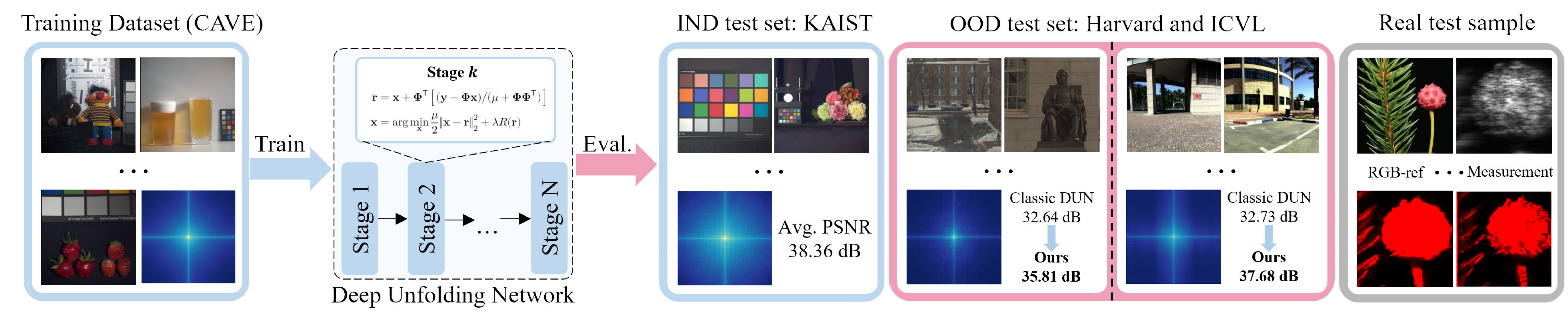}
    \caption{
    \textbf{Evaluating the cross-domain generalization of classic DUN.}
    Images with blue backgrounds represent the average log-spectrum of each dataset.
    The deep unfolding network is trained on the CAVE dataset during the slow learning stage.
    To evaluate cross-domain generalization, the model is tested on two out-of-distribution (OOD) datasets (Harvard and ICVL) and on real-world test samples, rather than on in-distribution (IND) datasets such as KAIST used in prior works.
    The average PSNR (dB) demonstrates that classic DUN performs well on IND test set but experiences a noticeable performance drop when applied to OOD test samples.
    }
    \label{fig:datasets}
\end{figure*}

In particular, deep unfolding frameworks (DUNs) unroll iterative optimization into trainable layers, thereby integrating model-driven regularization with data-driven learning \cite{meng2020gap,huang2021deep,Zhang_2024_CVPR_dpu,dauhst,Dong_2023_CVPR_rdluf,S2Trans_TPAMI2025}. 
However, these methods depend on extensive pre-training over large, paired datasets—an impractical requirement given the scarcity and acquisition cost of real HSI data, reflecting only the slow component of learning. 
Consequently, this heavy reliance leads to their poor performance on out-of-distribution (OOD) datasets, as illustrated in Fig. \ref{fig:datasets}.
Moreover, to achieve state-of-the-art performance, deep unfolding approaches often employ a large number of sequential unfolding blocks; each block must process the full high-dimensional hyperspectral volume, which greatly increases computational and memory burdens. This complexity not only complicates training and inference but also makes them impractical for real-world applications and deployment on resource-constrained or edge devices. 

Per-sample adaptation—analogous to human fast learning—is essential for robust SCI reconstruction under scarce HSI data. While recent advances such as self-supervised fine-tuning and test-time adaptation enable rapid calibration, applying them directly to deep unfolding networks proves problematic: full-network fine-tuning destabilizes sequential stages and linear probes cannot accommodate layer-wise inference. 
\cite{DEIL_TPAMI2022} pioneered external-internal learning for CASSI by pre-training on external data and then updating the entire
network for each test measurement using imaging consistency. However, full-network fine-tuning destabilizes sequential stages in
deep unfolding architectures.
SAH-SCI \cite{liu2024sahsci} supports general self-supervised tuning; however, its performance improvements are marginal, and it lacks runtime adaptation.
Recent test-time adaptation (TTA) methods \cite{eccv2024_tta, Qin_2023_CVPR, quan2022dual} exhibit sluggish inference even in basic RGB image compressive sensing scenarios, precluding their application to high-dimensional HSIs under practical time constraints.


To overcome these challenges, we introduce SlowFast-SCI, the first test-time adaptation-driven deep unfolding framework for efficient, self-adaptive spectral reconstruction. In the slow learning stage, we pre-train a deep unfolding backbone that fuses model-based priors with learnable parameters to acquire robust, general reconstruction capabilities. 
Furthermore, DUNs typically rely on numerous sequential stages to ensure high reconstruction fidelity, but this depth comes at the cost of heavy computation and slow inference. 
To address this, we apply an imaging mechanism-guided fast unfolding distillation step that compresses the pre‐trained backbone into a streamlined “fast unfolding” variant with significantly reduced complexity.
In the fast learning stage, lightweight Fast TTA modules (FAST-TTAMs) are inserted into each fast unfolding block and self-supervisedly fine-tuned at test time, instantly aligning the model to new measurements while preserving the frozen backbone. By uniting offline robustness with on-the-fly adaptation, SlowFast-SCI achieves superior cross-domain generalization without additional labels.
With the backbone frozen, this design delivers swift test‐time tuning and efficient inference, mirroring the human ability to rapidly leverage distilled knowledge for new experiences. 
We summarize our contributions as follows:
\begin{itemize}
  \item We propose a unified dual-speed learning framework, SlowFast-SCI, which augments any deep unfolding network with offline pre-training for robust reconstruction and lightweight on-the-fly adaptation to new spectral configurations. 
  To the best of our knowledge, SlowFast-SCI is the first parameter-efficient, stage-wise test-time adaptation framework tailored to deep-unfolding CASSI reconstruction, where a frozen backbone is adapted through lightweight per-stage modules.
  \item A imaging-guided fast unfolding distillation pipeline that compresses the pre-trained backbone into a fast unfolding variant, reducing parameter count and FLOPs by over 70\% while preserving adaptability.
  \item We develop a lightweight self-supervised adaptation module embedded in each fast unfolding stage to enable rapid per-sample calibration at inference.
  \item Evaluation on standard Harvard, ICVL and real CASSI measurements, showing up to 5.79 dB PSNR gain on out‐of‐distribution (OOD) setups and a $4 \times$ speedup in adaptation time.
\end{itemize}

\section{Related Work}

\begin{figure*}[t]
    \centering
    \includegraphics[width=1\linewidth]{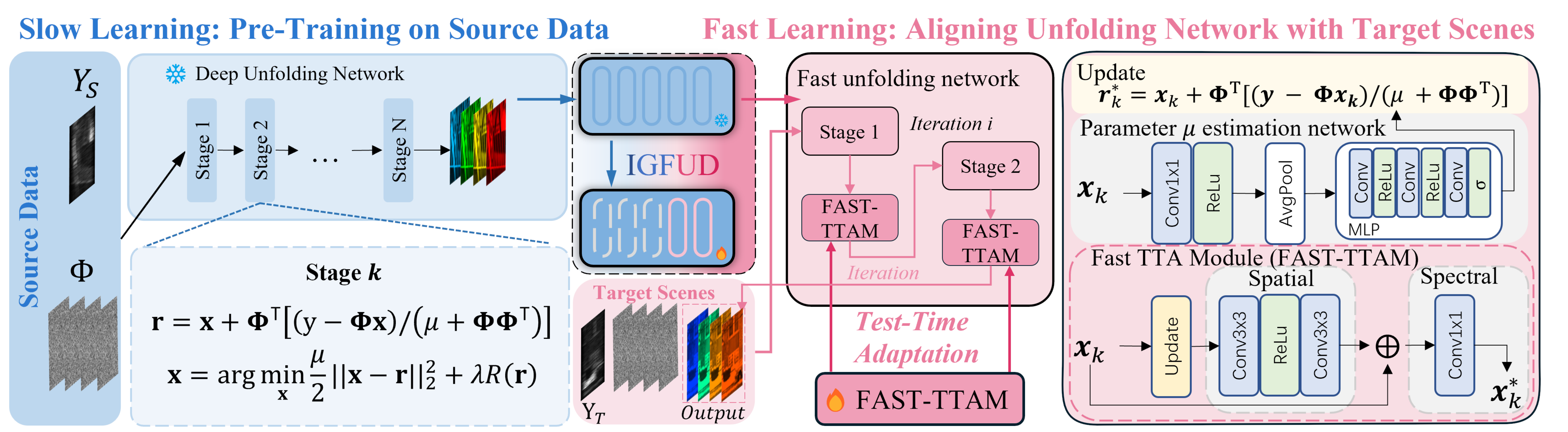}
    \caption{
    \textbf{Illustration of SlowFast-SCI.} 
    The left side depicts the \textcolor{cyan}{slow learning}\ process, pre-training on the source data with a deep unfolding network.
    IGFUD distills the knowledge of the deep unfolding backbone into a compact fast unfolding network (FastUN).
    The right side shows the \textcolor{magenta}{fast learning}\ process, where FAST-TTAM rapidly adapts the FastUN to align the model with target scenes at inference time.
    The far right illustrates the details of FAST-TTAM, which takes both spatial and spectral information into account during fast learning.
    }
    \label{fig:pipline}
\end{figure*}

\textbf{Hyperspectral Image Reconstruction.}
Related work comprises two main streams. 
Traditional optimization-based methods employ hand-crafted priors (\textit{e.g.}, sparsity \cite{wagadarikar2008single_cassi_sparsity,kittle2010multiframe_sparsity}, total variation \cite{rudin1992nonlinear_tv,wang2015dual_tv}, and low-rank constraints \cite{liu2018rank_lowrank,zhang2019computational_lowrank,DELTA_TPAMI2026}) 
to iteratively solve the inverse problem; while interpretable, they are limited by prior expressiveness and suffer from slow convergence. 
Deep learning and hybrid approaches accelerate SCI reconstruction \cite{mst,cst,dauhst,hdnet,Dong_2023_CVPR_rdluf,Zhang_2024_CVPR_dpu,S2Trans_TPAMI2025}, including end-to-end networks, plug-and-play frameworks, and deep unfolding architectures (\textit{e.g.}, S$^2$-Transformer \cite{S2Trans_TPAMI2025}, DGSM \cite{DGSM_TPAMI2023}, DAUHST \cite{dauhst}, RDLUF \cite{Dong_2023_CVPR_rdluf}, DPU \cite{Zhang_2024_CVPR_dpu}). 
Although these learning-based methods deliver state-of-the-art accuracy and speed, they rely heavily on large paired datasets, making them brittle under optical variations or spectral distribution shifts at test time. In contrast, our work introduces a complementary paradigm: a self-supervised, test-time adaptation framework that enables rapid per-sample calibration without additional labeled data, thus improving robustness across unseen scenes.

\noindent \textbf{Test-Time Adaptation.}
TTA adapts a pre-trained model during inference to improve performance without labeled test samples \cite{sun2020test_tta, darestani2022test_tta}.
In hyperspectral imaging, DEIL \cite{DEIL_TPAMI2022} introduced external-internal learning for CASSI reconstruction, updating the
full network per measurement via imaging consistency—but at high computational cost.
In image reconstruction, TTA fine-tunes models at test time \cite{quan2022dual,Qin_2023_CVPR,eccv2024_tta}. AdaptNet \cite{eccv2024_tta} integrates self-supervised test time training into deep unfolding networks, but its considerable depth leads to slow adaptation. In contrast, we propose a lightweight adaptation module and a concise imaging-guided fast unfolding distillation pipeline, enabling efficient test-time adaptation via adapter architectures.

\noindent\textbf{Model Distillation.}
Model distillation~\cite{bucilua2006model_distill, hinton2015distilling} is a widely adopted method for transferring knowledge from a large teacher to a smaller student network. While initial methods used soft-label supervision, subsequent work extended distillation via intermediate feature matching~\cite{romero2014fitnets} and attention transfer~\cite{Zagoruyko2017AT}, and applied it to self-supervised learning~\cite{caron2021emerging} and TTA~\cite{wang2021tent}. Existing distillation primarily targets semantic tasks, leaving inverse problems---especially SCI-based image reconstruction---largely unexplored. We address this by introducing a structural-knowledge distillation framework that transfers imaging priors from a pre-trained backbone into a lightweight, fast-unfolding student model. The student inherits the teacher’s robustness while remaining efficient and adaptable during test-time.

\section{Imaging Model and Deep-Unfolding Preliminaries}
\label{prelim}

\begin{figure}[t]
    \centering
    \includegraphics[width=0.98\linewidth]{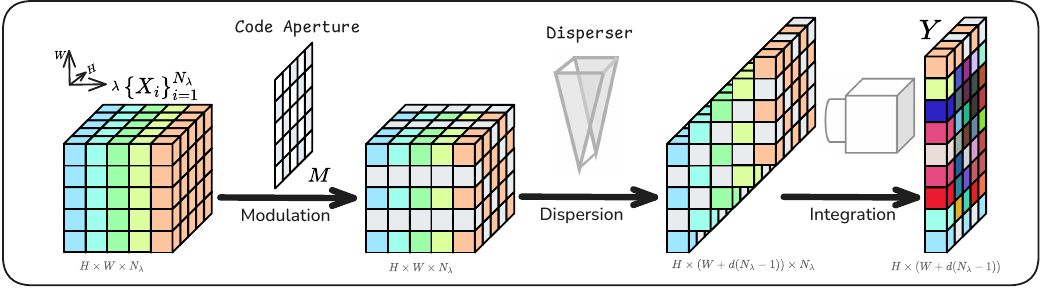}
    \caption{The schematic of CASSI system.}
    \label{fig:cassi}
\end{figure}

\textbf{Mathematical Model of CASSI.}
SCI is a snapshot modality that captures a full hyperspectral data cube in a single coded measurement by jointly modulating and integrating across all spectral bands. 
As illustrated in Fig.~\ref{fig:cassi}, the CASSI implements SCI as follows.
Let a hyperspectral patch consist of $N_{\lambda}$ bands $\{X_i\}_{i=1}^{N_{\lambda}}$, each $X_i\in\mathbb{R}^{H\times W}$. Each band is first modulated by a known binary mask $M$, then the modulated frames are shifted along the detector by $d(i-1)$ pixels and summed to yield the compressed measurement:
\begin{equation*}
Y(u,v) = \sum_{i=1}^{N_{\lambda}} M \odot X_i\bigl(u,\,v + d(i-1)\bigr) + N(u,v),
\end{equation*}
where $\odot$ denotes the Hadamard product, $N$ is additive sensor noise, and $(u,v)$ index spatial coordinates. By vectorizing the 3D hyperspectral cube into $\mathbf{x}\in\mathbb{R}^{nN_{\lambda}}$ and the 2D measurement into $\mathbf{y}\in\mathbb{R}^{n}$ with $n = H(W + d(N_{\lambda}-1))$, the acquisition model becomes $\mathbf{y} = \Phi\mathbf{x} + \mathbf{n}$, 
where $\Phi\in\mathbb{R}^{n\times nN_{\lambda}}$ encodes the combined masking and shifting operations.

\textbf{MAP Formulation.}
Deep unfolding networks formulate SCI reconstruction as a Bayesian
problem under a unified MAP framework:
\begin{equation}
    \hat{\mathbf{x}}_{\mathit{map}} = \arg\min_{\mathbf{x}} \log P(\mathbf{y}|\mathbf{x}) + \log P(\mathbf{x}).
    \label{MAP}
\end{equation}
The original signal can be estimated by minimizing the energy function
\begin{equation}
     \hat{\mathbf{x}} = \arg\min_{\mathbf{x}} \tfrac{1}{2}\|\mathbf{y} - \Phi\mathbf{x}\|_2^2 + \lambda R(\mathbf{x}),
     \label{energy function}
\end{equation}
where the first term is the data fidelity, $R(\mathbf{x})$ is the
image prior, and $\lambda$ is a trade-off hyperparameter.

\textbf{HQS Iterative Scheme.}
Among many solutions for Eq.~(\ref{energy function}), the HQS method
formulates the problem as the iterative scheme:
\begin{align}
    \mathbf{r}_{k+1} &= \mathbf{x}_{k} + \Phi^{\mathsf{T}} \left[ (\mathbf{y} - \Phi \mathbf{x}_{k})/(\mu + \Phi\Phi^{\mathsf{T}}) \right] ,
    \label{r}
    \\
    \mathbf{x}_{k+1} &= \arg\min_{\mathbf{x}} \tfrac{\mu}{2} \| \mathbf{x} - \mathbf{r}_{k+1} \|_2^2 + \lambda R(\mathbf{x}),
    \label{x}
\end{align}
where $\mu$ is a penalty parameter. Based
on this iterative scheme, one can construct a deep unfolding network
with many stages. Within each stage, the $\mathbf{x}$-subproblem can
be regarded as a denoising problem: predicting the original HSI
signals $\mathbf{x}$ from their noisy counterparts
$\mathbf{r} = \mathbf{x} + \epsilon$, solved by a learnable denoising
network.

\section{Method}

SlowFast-SCI tackles two complementary challenges in SCI reconstruction with dual-speed learning. In the \textcolor{cyan}{slow learning}\ stage (Sec. \ref{sec:slow-learning}), we overcome the ill-posed inverse problem and scarce HSI training data by pre-training a physics-guided unfolding backbone that fuses model priors with learnable parameters to learn robust, general reconstruction capabilities. 
In the \textcolor{magenta}{fast learning}\ stage (Sec. \ref{sec:fast-learning}), we address domain shifts and per-sample variability by inserting lightweight spectral adapters into each unfolding block and self-supervisedly fine-tuning only these adapters at test time—instantly calibrating to unseen measurements while keeping the backbone frozen. Furthermore, we propose an imaging-guided fast unfolding distillation framework that condenses the hybrid architecture into a compact “fast unfolding” network, augmented with tailored spectral adaptation blocks. This design enables efficient, label-free reconstruction across domains, mirroring the rapid application of distilled expertise in human learning.

\subsection{Slow Learning} \label{sec:slow-learning}
\begin{figure}[h]
    \centering
    \includegraphics[width=\columnwidth]
        {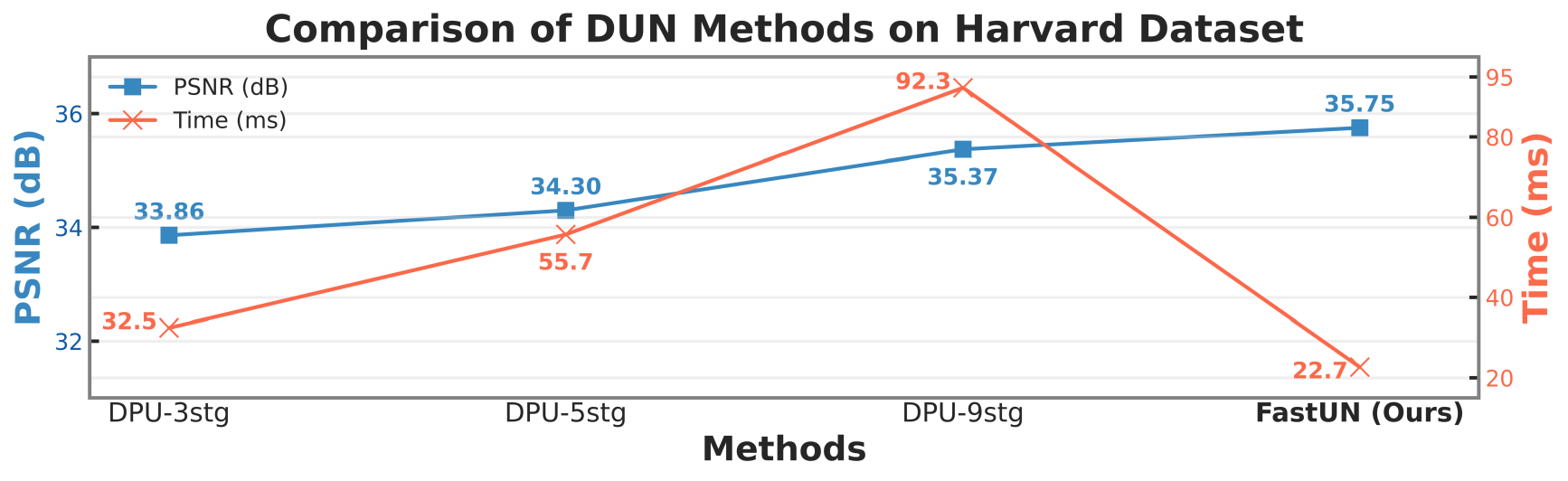}

    \caption{
    Comparison of DUN methods on 10 scenes from the \texttt{Harvard}
    dataset, reporting PSNR (dB) and per-scene inference time (ms).
    For DPU, results with different stage counts are shown.
    }
    \label{fig:end_dun}
\end{figure}

In the slow learning stage, we pre-train a physics-guided unfolding backbone that integrates analytical priors with learnable parameters to address the ill-posed nature of spectral reconstruction and the scarcity of hyperspectral data. By explicitly embedding the underlying imaging physics into the network design, the unfolding backbone acquires a robust and generalizable reconstruction 
capability even under limited supervision. 
\begin{wraptable}{l}{0.42\linewidth} 
\vspace{-3mm}
\centering
\captionsetup{justification=justified, width=0.22\textwidth, font=scriptsize}
\caption{\scriptsize Comparison between the end-to-end (E2E) network and the deep unfolding network (DUN), both employing the same Transformer backbone.}
\setlength\tabcolsep{2pt}
\scriptsize
\begin{tabular}{lcc}
Method & Category & PSNR(dB) \\ 
\hline
MST \cite{mst} & E2E & 31.18 \\ 
DAUHST \cite{dauhst} & DUN & 32.64 \\ 
\end{tabular}
\label{tab:end_dun}
\vspace{-4mm}
\end{wraptable}
As illustrated in Tab.~\ref{tab:end_dun}, when employing 
the same backbone architecture (e.g., Transformer), the DUN generally achieves superior performance compared to its end-to-end counterpart.
Furthermore, DUNs operate in a multi-stage iterative manner (e.g., 3, 5, or 9 stages), where increasing the number of stages progressively refines reconstruction performance at the cost of higher inference latency, as shown in Fig.~\ref{fig:end_dun}.
Building upon this foundation, the slow learning stage yields a 9-stage deep unfolding backbone with strong reconstruction accuracy. 
However, directly deploying this full-depth model in the fast learning stage is suboptimal; its massive parameter space is overly burdensome for single-scene adaptation, which hinders effective few-step parameter updates and impedes rapid adaptation to novel scenes. As illustrated in Tab. \ref{tab:re_main}, previous methods \cite{eccv2024_tta} follow this paradigm, resulting in prohibitive inference latency and compromised robustness when encountering OOD data. Additionally, training a shallow unfolding network directly on source data is equally ineffective due to its limited representational capacity.
To address these limitations, we propose Imaging-Guided Fast Unfolding Distillation (IGFUD), which bridges the gap between high-fidelity reconstruction and real-time adaptability by distilling the knowledge of the deep unfolding backbone into a compact fast unfolding network, thereby reducing model complexity while enabling unsupervised adaptation to unseen imaging conditions.

\begin{table}[t]
    \centering
    \caption{
    Quantitative comparison of performance and efficiency between our
    proposed method and existing approaches on the Harvard dataset (OOD).
    All adaptation methods share the same self-supervised loss, number
    of iterations, and trainable-adapter accounting for a fair comparison.
    }
    \label{tab:re_main}

    \resizebox{\columnwidth}{!}{%
        \begin{tabular}{c c c c c c}
        \hline
        Method & PSNR (dB) & SSIM & Params (M) & FLOPs (G) & Time (s) \\
        \hline
        DPU-9stg
            & 35.37 & 0.868 & 2.85 & 49.26 & -- \\

        DPU-9stg + Adapters (Previous)
            & \cellcolor{rouse}35.00
            & \cellcolor{rouse}0.851
            & 3.16 & 67.92 & $\sim$7.93 \\

        Directly Trained 2-stage Shallow UN
            & \cellcolor{rouse}33.54
            & \cellcolor{rouse}0.820
            & 0.64 & 10.86 & -- \\

        2-stage FastUN (Ours)
            & \color{eccvblue}35.75
            & \color{eccvblue}0.876
            & 0.64 & 10.86 & -- \\

        SlowFast-SCI$^2$ (Ours)
            & \color{red}40.33
            & \color{red}0.931
            & 0.70 & 15.17 & $\sim$1.82 \\
        \hline
        \end{tabular}%
    }
\end{table}

\textbf{Imaging-Guided Fast Unfolding Distillation.}
The fast unfolding distillation process is shown in Fig.~\ref{fig:pipline}.
During this process, the pre-trained model, with its parameters frozen, serves as the slow unfolding network (SlowUN), while the two-stage model with the same architecture as the pre-trained model is used as fast unfolding network (FastUN).
The dataset (measurements: $\{\mathrm{\mathbf{y}}_i\}_{i=1}^{N} \in \mathbb{R}^{H\times(W + d\times(N_{\lambda} - 1)}$) utilized for IGFUD is strictly unseen by the SlowUN during the supervised slow learning stage, as shown in Fig. \ref{fig:datasets}. 
Furthermore, we partition these measurements into two distinct subsets: $\{\mathrm{\mathbf{y}}_i\}_{i=1}^{N_{d}} \in \mathbb{R}^{H\times(W + d\times(N_{\lambda} - 1))}$ and $\{\mathrm{\mathbf{y}}_i\}_{i=1}^{N_{t}} \in \mathbb{R}^{H\times(W + d\times(N_{\lambda} - 1))}$ ($N = N_{d} + N_{t}$), where the latter is exclusively utilized for the ultimate test-time adaptation phase.
Firstly, mask $\Phi$ and measurements $\{\mathrm{\mathbf{y}}_i\}_{i=1}^{N_{d}} \in \mathbb{R}^{H\times(W + d\times(N_{\lambda} - 1)}$ are entered into the SlowUN to obtain the results $\{\mathrm{\mathbf{X}}_{S,i}\}_{i=1}^{N_d} \in \mathbb{R}^{H\times W \times N_{\lambda}}$, which will be used as a label to supervise the training of FastUN. 
Then mask $\Phi$ and measurements $\{\mathrm{\mathbf{y}}_i\}_{i=1}^{N_d}$ are input into FastUN to
obtain the results $\{\mathrm{\mathbf{X}}_{F, i}\}_{i=1}^{N_d}$. 
We update FastUN by minimizing the following loss:
\begin{equation}
    \mathscr{L}_{IGFUD} = \frac{1}{N_d} \sum_{i=1}^{N_d} \|\mathrm{\mathbf{X}}_{F,i} - \mathrm{\mathbf{X}}_{S,i} \|_F^2.
\end{equation}
The inference speed of the two-stage FastUN obtained by IGFUD is much faster than the original pre-trained model, and its performance on test data is comparable to that of the original pre-trained model, as shown in Fig. \ref{fig:end_dun}.
Notably, relying solely on measurements to train a shallow unfolding network often yields unstable training and poor feature representation due to missing supervision. By adopting IGFUD, however, we successfully obtain a robust and efficient FastUN.

\begin{figure*}[t]
    \centering
    \includegraphics[width=1\linewidth]{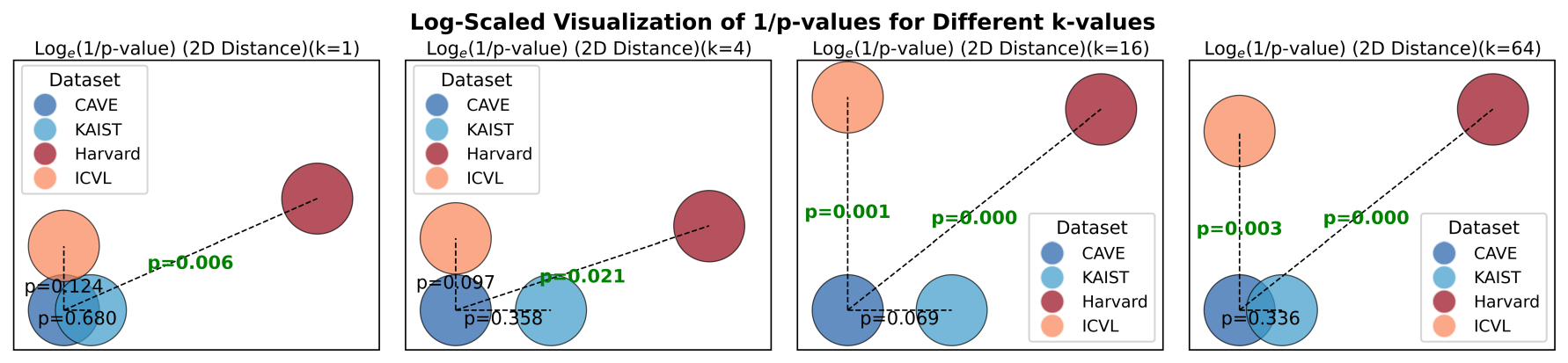}
    \caption{
    \textbf{Log-scaled visualization of 1/p-values for different k-values.}
    The p-values obtained by performing a permutation test ($n = 1000$) on the
    kernel matrices derived from the Maximum Mean Discrepancy \cite{rabanser2019failing} between CAVE (the pre-training dataset) and the other three datasets (KAIST, Harvard, and
    ICVL). 
    Each image (originally of size $256 \times 256 \times 28$) is processed using average pooling with different window sizes $k$. This step aims to attenuate the spatial features, making the spectral dimension more prominent for distribution detection. \textbf{\textcolor{green}{Green bold}} values indicate statistically significant differences ($p < 0.05$) between the two data distributions.
    }
    \label{fig:mmd}
    \vspace{-2mm}
\end{figure*}

\subsection{Fast Learning} \label{sec:fast-learning}

Our slow-learning phase (Sec. \ref{sec:slow-learning}) endows the deep unfolding network with robust physics-informed priors and learnable parameters (\textit{i.e.}, the denoising network).
However, when encountering a new spectral domain, naively fine-tuning all unfolding blocks requires nearly the same amount of data and computation as the original training process.
Therefore, we aim to develop a more efficient strategy to adapt the model learned during the slow-learning stage.

Eqs. (\ref{r}) and (\ref{x}) formally describe the model obtained from the slow-learning phase, where Eq. (\ref{r}) provides a closed-form solution that is independent of the data distribution.
Accordingly, the component of the model that is highly sensitive to the data distribution resides in Eq. (\ref{x})—more precisely, in $R(\mathbf{r})$, which is modeled by a learnable denoising network during slow learning.

By roughly analyzing the statistical properties as a form of Bayesian inference of this denoising network, we design the Wiener-Filter-Inspired Adapter Modules, which enable the model to adapt efficiently to test samples whose distributions are similar to that of the training data.
However, as illustrated in Fig. \ref{fig:mmd}, the distributions of the two test datasets differ substantially from that of the pre-training dataset.
To bridge this distribution gap and enhance the model’s adaptability, we propose Self-Adaptation-Driven Fast-Learning Unfolding.

\subsubsection{Wiener-Filter-Inspired Adapter Modules.}\label{sec:wiener}
How to interpret the statistical properties of general CNNs within the framework of Bayesian inference remains an open question.
For clarity of discussion, we denote the optimal denoising filter—\textit{i.e.}, the Wiener filter—for each spectral channel $\lambda_i$ corresponding to a training dataset $\mathrm{\mathbf{x}}$ and a specific test sample $\tilde{\mathrm{\mathbf{x}}}$ as follows, respectively:
\begin{equation}
    \widehat{\mathcal{H}}_{w, \lambda_i}[k_1, k_2] = \frac{\mathbb{E}_{\mathbf{x}}\{|\widehat{\mathbf{x}}_{\lambda_i}[k_1, k_2]|^2\}}{\mathbb{E}_{\mathbf{x}}\{|\widehat{\mathbf{x}}_{\lambda_i}[k_1, k_2]|^2\} + \mathbb{E}_{\mathbf{\epsilon}}\{|\widehat{\mathbf{\epsilon}}_{\lambda_i}[k_1, k_2]|^2\}},
    \label{wiener}
\end{equation}
\begin{equation}
    \widehat{\widetilde{\mathcal{H}}}_{w,\mathscr{\widetilde{\mathbf{x}}}, \lambda_i}[k_1,k_2] = \frac{|\widehat{\widetilde{\mathbf{x}}}_{\lambda_i}[k_1,k_2]|^2}{|\widehat{\widetilde{\mathbf{x}}}_{\lambda_i}[k_1,k_2]|^2 + \mathbb{E}_{\widehat{\mathbf{\epsilon}}}\{|\widehat{\widetilde{\mathbf{\epsilon}}}_{\lambda_i}[k_1,k_2]|^2\}}.
    \label{targetwiener}
\end{equation}
Here, $[k_1, k_2]$ denotes the spatial frequency components obtained via the discrete Fourier transform (DFT).
For simplicity, inter-channel spectral correlations are neglected in this formulation.
Detailed derivations of Eqs.~(\ref{wiener}) and~(\ref{targetwiener}), yielding the optimal Wiener filter by minimizing the frequency-domain MSE, are provided below.
If the test sample’s spectrum is close to the training data distribution’s expectation,
the Wiener solution for a specific target scene (Eq. (\ref{targetwiener})) only differs modestly from the source Wiener solution (Eq. (\ref{wiener})).
Thus, a compact, supplementary lightweight adapter can bridge this gap, aligning the pre-trained source filter with the target‐scene Wiener filter.

\begin{proof}
Let $\mathcal{D}(\cdot, \omega)$ denote a denoising network with parameters $\omega$. 
Then the Eq. (5) is transformed into the following form:
\begin{equation}
    \mathrm{\mathbf{x}}_{k+1} = \mathcal{D}(\mathrm{\mathbf{r}}_{k+1}, \omega_{k+1}).
\end{equation}
Training a denoising network $\mathcal{D}(\cdot, \omega)$ with parameters $\omega$ to predict the original HSI signals $\mathbf{x}$ from their noisy counterparts $\mathbf{r} = \mathbf{x} + \mathbf{\epsilon}$, \textit{i.e.}, to minimize prediction error means minimizing the mean squared error (MSE) by updating the parameters $\omega$:
\begin{equation}
    \min_{\theta} \mathbb{E}_{\mathbf{x} \sim p(\mathbf{x}), \mathbf{\epsilon} \sim p(\mathbf{\epsilon})} \{ \| \mathcal{D}(\mathrm{\mathbf{r}}, \omega) - \mathbf{x} \|_F^2 \}.
\end{equation}

How to understand the statistical properties of general CNNs as a form of Bayesian inference is an open question. For the sake of discussion, we ignore the correlation between spectral channels and consider a CNN consisting of only linear convolution layers, which does not contain nonlinear activation functions and bias terms, and its expression is:
\begin{equation}
    \mathcal{D}(\mathrm{\mathbf{r}}, \omega) = \mathcal{H} \otimes \mathrm{\mathbf{r}},
\end{equation}
where $\otimes$ represents the convolution operator and $\mathcal{H}$ is the filter of $N_{\lambda}$ channels. 

The optimal filter, \textit{Wiener filter}, aims at minimizing the Bayesian risk:

\begin{equation}
    \mathcal{H}_w := \arg\min_{\mathcal{H}}\mathbb{E}_{\mathrm{\mathbf{x}} \sim p(\mathrm{\mathbf{x}}), \mathbf{\epsilon}\sim p(\mathbf{\epsilon})} \{ \| \mathcal{H} \otimes \mathrm{\mathbf{r}} - \mathrm{\mathbf{x}} \|_F^2 \}.
\end{equation}

The Wiener filter is defined in the frequency domain using the discrete Fourier transform (DFT), denoted by $\widehat{(\cdot)}$.
Since $\widehat{(\cdot)}$ is a unitary transform, it holds for each frequency component $[k_1, k_2, \lambda_i]$ of each channel $\lambda_i$:
\begin{equation}
\begin{split}
    \widehat{\mathcal{H}}_w :&= \arg\min_{\widehat{\mathcal{H}}}\mathbb{E} \{ \| \widehat{\mathcal{H}} \odot \widehat{\mathrm{\mathbf{r}}} - \widehat{\mathrm{\mathbf{x}}} \|_F^2 \} \\
    &= \arg\min_{\widehat{\mathcal{H}}}\sum_{\lambda_i} \sum_{k_1} \sum_{k_2} e(k_1, k_2, \lambda_i),
\end{split}
\end{equation}
where $e(k_1, k_2, \lambda_i) = \mathbb{E} \{ |\widehat{\mathcal{H}}[k_1, k_2, \lambda_i] \widehat{\mathrm{\mathbf{r}}}[k_1, k_2, \lambda_i] - \widehat{\mathrm{\mathbf{x}}}[k_1, k_2, \lambda_i]|^2 \} $ and $\odot$ denotes the element-wise multiplication.
Expanding $e(k_1, k_2, \lambda_i)$ further, we can get the following

\begin{equation}
\resizebox{1.0\columnwidth}{!}{$\displaystyle
\begin{aligned}
e(k_1, k_2, \lambda_i)
&= \mathbb{E}_{\mathrm{\mathbf{x}}, \mathbf{\epsilon}} \{ |\widehat{\mathcal{H}}[k_1, k_2, \lambda_i](\widehat{\mathrm{\mathbf{x}}}[k_1, k_2, \lambda_i] + \widehat{\epsilon}[k_1, k_2, \lambda_i])  - \widehat{\mathrm{\mathbf{x}}}[k_1, k_2, \lambda_i]|^2 \} \\
&= \mathbb{E}_{\mathrm{\mathbf{x}}, \mathbf{\epsilon}} \{ | \widehat{\mathrm{\mathbf{x}}}[k_1, k_2, \lambda_i] (\widehat{\mathcal{H}}[k_1, k_2, \lambda_i] - 1) + \widehat{\mathcal{H}}[k_1, k_2, \lambda_i] \widehat{\epsilon}[k_1, k_2, \lambda_i] |^2 \} \\
&= (\widehat{\mathcal{H}}[k_1, k_2, \lambda_i] - 1)(\widehat{\mathcal{H}}[k_1, k_2, \lambda_i] - 1)^* \mathbb{E}_{\mathrm{\mathbf{x}}} \{ |\widehat{\mathrm{\mathbf{x}}}[k_1, k_2, \lambda_i]|^2 \} \\
&\quad - (\widehat{\mathcal{H}}[k_1, k_2, \lambda_i] - 1)\widehat{\mathcal{H}}^*[k_1, k_2, \lambda_i] \mathbb{E}_{\mathrm{\mathbf{x}}, \mathbf{\epsilon}} \{ \widehat{\mathrm{\mathbf{x}}}[k_1, k_2, \lambda_i] \widehat{\epsilon}^*[k_1, k_2, \lambda_i] \} \\
&\quad - (\widehat{\mathcal{H}}[k_1, k_2, \lambda_i] - 1)^* \widehat{\mathcal{H}}[k_1, k_2, \lambda_i] \mathbb{E}_{\mathrm{\mathbf{x}}, \mathbf{\epsilon}} \{ \widehat{\mathrm{\mathbf{x}}}^*[k_1, k_2, \lambda_i] \widehat{\epsilon}[k_1, k_2, \lambda_i] \} \\
&\quad + \widehat{\mathcal{H}}[k_1, k_2, \lambda_i] \widehat{\mathcal{H}}^*[k_1, k_2, \lambda_i] \mathbb{E}_{\mathbf{\epsilon}} \{ |\widehat{\epsilon}[k_1, k_2, \lambda_i]|^2 \},
\end{aligned}
$}
\end{equation}

where $*$ denotes the complex conjugation. Based on the assumption that $\mathrm{\mathbf{x}}$ and $\mathbf{\epsilon}$ are
independent and $\mathbb{E}[\mathbf{\epsilon}] = 0$, we have
$\mathbb{E}_{\mathbf{x}, \mathbf{\epsilon}} \{ \widehat{\mathrm{\mathbf{x}}}[k_1, k_2, \lambda_i] \widehat{\epsilon}^*[k_1, k_2, \lambda_i] \} = \mathbb{E}_{\mathbf{x}, \mathbf{\epsilon}} \{ \widehat{\mathrm{\mathbf{x}}}^*[k_1, k_2, \lambda_i] \widehat{\epsilon}[k_1, k_2, \lambda_i] \} = 0$. 

Thus
\begin{equation}
\resizebox{1.0\columnwidth}{!}{$\displaystyle
\begin{aligned}
    e(k_1, k_2, \lambda_i) &= (\widehat{\mathcal{H}}[k_1, k_2, \lambda_i] - 1)(\widehat{\mathcal{H}}[k_1, k_2, \lambda_i] - 1)^*  \mathbb{E}_{\mathrm{\mathbf{x}}} \{ |\widehat{\mathrm{\mathbf{x}}}[k_1, k_2, \lambda_i]|^2 \} \\
    & + \widehat{\mathcal{H}}[k_1, k_2, \lambda_i] \widehat{\mathcal{H}}^*[k_1, k_2, \lambda_i] 
    \mathbb{E}_{\mathbf{\epsilon}} \{ |\widehat{\epsilon}[k_1, k_2, \lambda_i]|^2 \}.
\end{aligned}
$}
\end{equation}
Based on the assumption that the real and imaginary part of $\widehat{\mathcal{H}}[k_1, k_2, \lambda_i]$ are independent, 
the complex variable $\widehat{\mathcal{H}}[k_1, k_2, \lambda_i]$ and complex conjugate variable $\widehat{\mathcal{H}}^*[k_1, k_2, \lambda_i]$ are two independent variables.
Then, we calculate the partial derivative with respect to $\widehat{\mathcal{H}}[k_1, k_2, \lambda_i]$ and let
\begin{equation}
\begin{split}
    \frac{\partial e(k_1, k_2, \lambda_i)}{\partial \widehat{\mathcal{H}}[k_1, k_2, \lambda_i]} &=
    (\widehat{\mathcal{H}}[k_1, k_2, \lambda_i] - 1)^* \mathbb{E}_{\mathrm{\mathbf{x}}} \{ |\widehat{\mathrm{\mathbf{x}}}[k_1, k_2, \lambda_i]|^2 \} \\
    &+ \widehat{\mathcal{H}}^*[k_1, k_2, \lambda_i] \mathbb{E}_{\mathbf{\epsilon}} \{ |\widehat{\epsilon}[k_1, k_2, \lambda_i]|^2 \}
    = 0 .
\end{split}
\end{equation}
Thus
\begin{equation}
    \widehat{\mathcal{H}}_\omega[k_1, k_2, \lambda_i] = 
    \frac{\mathbb{E}_{\mathrm{\mathbf{x}}} \{ |\widehat{\mathrm{\mathbf{x}}}[k_1, k_2, \lambda_i]|^2 \}}
     {\mathbb{E}_{\mathrm{\mathbf{x}}} \{ |\widehat{\mathrm{\mathbf{x}}}[k_1, k_2, \lambda_i]|^2 \}
     + \mathbb{E}_{\mathbf{\epsilon}} \{ |\widehat{\epsilon}[k_1, k_2, \lambda_i]|^2 \}}.
\end{equation}
\end{proof}


\subsubsection{Self-Adaptation-Driven Fast-Learning Unfolding.} \label{sec:TTA}

\textbf{Adapter-Enhanced Unfolding Network Architecture.}
Motivated by the analysis in Sec. \ref{sec:wiener}, we 
propose an adapter-enhanced unfolding network (AdaptUN).
AdaptUN is built upon our previously introduced FastUN (see Sec. \ref{sec:slow-learning}) by appending a lightweight module—namely, the fast test-time adaptation module (FAST-TTAM)—after each unfolding stage.
FAST-TTAM jointly leverages spatial and spectral cues to facilitate efficient test-time adaptation.
During the adaptation phase, all parameters of the original unfolding backbone are frozen, and only the parameters within FAST-TTAM are learnable.
The FAST-TTAM is a lightweight operator with one update step, one parameter $\mu$ estimation network, and two convolution blocks, 
the spatial convolution block and the spectral convolution block. 
The structure of parameter $\mu$ estimation network is shown in the upper right of Fig. \ref{fig:pipline}, which is used to estimate the parameter $\mu$.
The spatial convolution block consists of two convolution layers with $3\times3$ filter and one $\mathrm{ReLU}$ function, 
while the spectral convolution block only includes one convolution layer with $1\times1$ filter. 


\begin{figure}[t]
    \centering
    \includegraphics[width=1\linewidth]{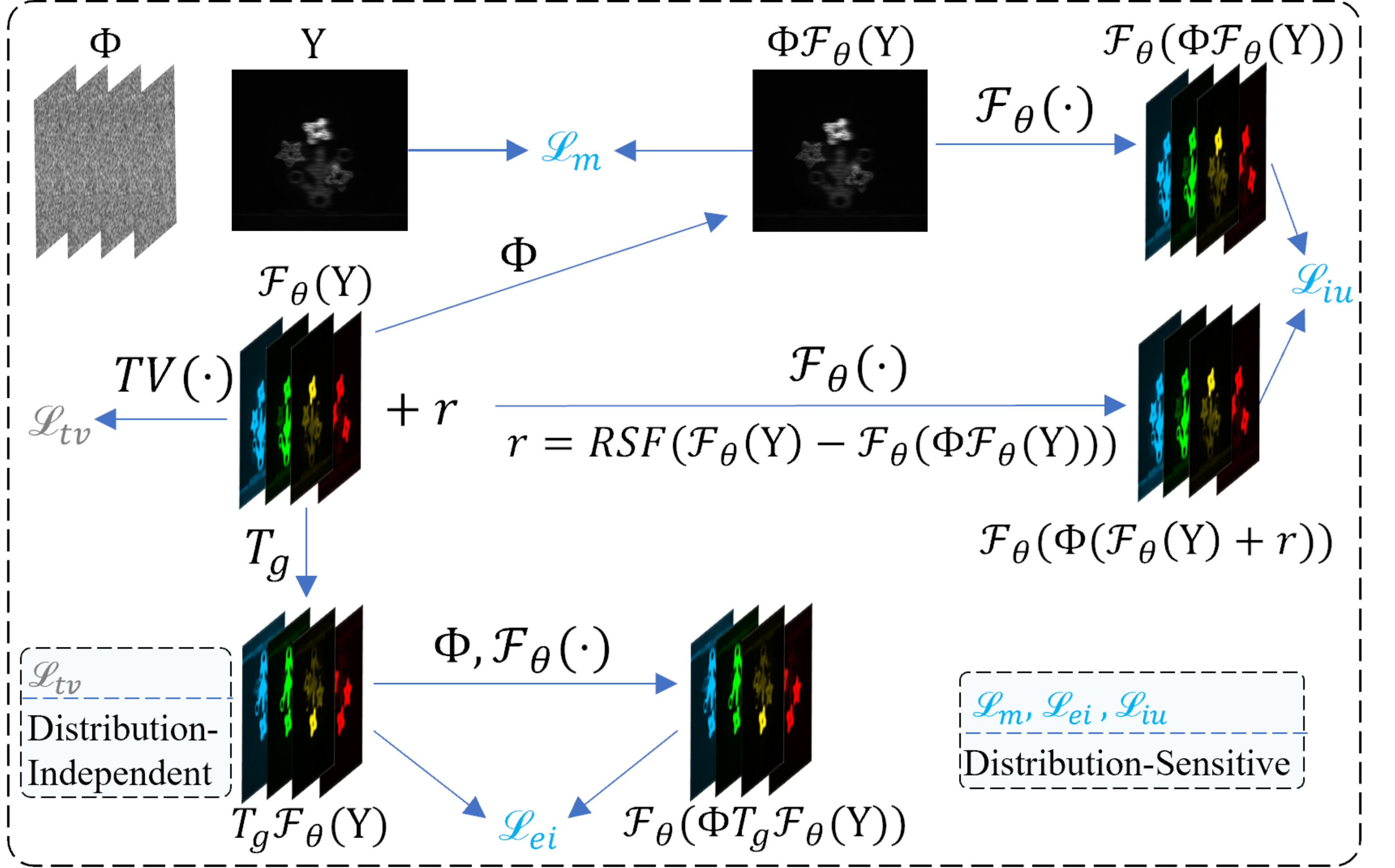}
    \caption{
    Illustration of four different self-supervised losses used in the fast learning stage.
    $\Phi$, $Y$, and $\mathcal{F}_{\theta}$ denote the mask, measurement, and model, respectively.
    $TV(\cdot)$ represents the total variation loss, and RSF stands for random sign-flipping sampling.
    $T_g$ denotes a set of geometric transformations (\textit{e.g.}, shifts, rotations, \textit{etc.}).
    The total variation loss $\mathscr{L}_{tv}$ is distribution-independent, serving as a regularization term to preserve image smoothness, whereas $\mathscr{L}_{m}$, $\mathscr{L}_{ei}$, and $\mathscr{L}_{iu}$ are distribution-sensitive.
    }
    \label{fig:loss}
\end{figure}

\textbf{Self-Supervised fine-tuning and Test-Time Adaptation.} 
Consider AdaptUN: $\mathcal{F}_{AdaptUN}(\cdot;\omega^*, \theta)$:
\begin{equation}
    \mathcal{F}_{AdaptUN}(\cdot;\omega^*, \theta): \mathrm{\mathbf{y}}, \mathbf{\Phi} \rightarrow{\mathrm{\mathbf{x}}},
    \label{distilled model}
\end{equation}
where $\omega^*$ denotes the parameters of FastUN frozen in self-supervised fine-tuning and test-time adaptation and 
$\theta$ denotes the parameters of the adapt layers to be updated, 
given the measurements $\{\mathrm{\mathbf{y}}_i\}_{i=1}^{N}$.
Learning the model solely from observations captured by a single incomplete mask is inherently impossible, as the ill-posed SCI inverse problem and the uninformative nullspace of the operator jointly result in infinite solutions and severe ambiguity within the solution space.
To overcome this, 
we update $\theta$ by minimizing the following self-supervised loss:
\begin{equation}
    \mathscr{L}_{sst} = \mathscr{L}_{m} + \omega_1\mathscr{L}_{ei} + \omega_2\mathscr{L}_{iu} + \omega_3\mathscr{L}_{tv}.
    \label{l_self}
\end{equation}
where $\mathscr{L}_m$, $\mathscr{L}_{ei}$, $\mathscr{L}_{iu}$, and $\mathscr{L}_{tv}$ denote the measurement consistency loss, equivariance imaging loss \cite{liu2024sahsci}, image uncertainty loss, and total variation loss \cite{rudin1992nonlinear_tv}, respectively.
The coefficients $\omega_1$, $\omega_2$, and $\omega_3$ represent the equilibrium constants.

\begin{table*}[t]
\def\arraystretch{1.1}
\setlength{\tabcolsep}{5pt}
\newcommand{\tabincell}[2]
{\begin{tabular}{@{}#1@{}}#2\end{tabular}}
\centering
\caption{Comparisons (PSNR (upper entry ineach cell), SSIM (lower entry in each cell), Params, and FLOPs) between SlowFast-SCI (SlowFast-SCI$^1$ and SlowFast-SCI$^2$ represent the slow-learning backbone we used were DAUHST and DPU, respectively.) and SOTA methods on 10 simulation scenes of \texttt{Harvard} dataset. 
}
\resizebox{0.99\textwidth}{!}
{
\centering
\begin{tabular}{ccccccccccccccc}
\toprule[0.1em]
\rowcolor{lightgray}
Method & Reference & Params (M) & FLOPs (G) & S1 & S2 & S3 & S4 & S5 & S6 & S7 & S8 & S9 & S10 & Avg \\
\midrule
TSA-Net \cite{meng2020endtsa_cassi} & ECCV 2020 & 42.20 & 91.58 &\tabincell{c}{30.54\\0.795}&\tabincell{c}{25.76\\0.557}&\tabincell{c}{30.27\\0.750}&\tabincell{c}{22.24\\0.422}&\tabincell{c}{36.71\\0.845}&\tabincell{c}{27.62\\0.820}&\tabincell{c}{27.51\\0.610}&\tabincell{c}{24.83\\0.644}&\tabincell{c}{27.23\\0.570}&\tabincell{c}
{22.79\\0.490}&\tabincell{c}{27.55\\0.650} \\

\midrule
MST \cite{mst} & CVPR 2022 & 1.92 & 25.60 &\tabincell{c}{32.67\\0.857}&\tabincell{c}{30.28\\0.699}&\tabincell{c}{31.20\\0.798}&\tabincell{c}{26.21\\0.515}&\tabincell{c}{38.63\\0.938}&\tabincell{c}{30.40\\0.837}&\tabincell{c}{32.80\\0.751}&\tabincell{c}{29.34\\0.791}&\tabincell{c}{32.40\\0.732}&\tabincell{c}{27.85\\0.670}&\tabincell{c}
{31.18\\0.759}   \\
\midrule
CST \cite{cst} & ECCV 2022 & 2.86 & 36.77 &\tabincell{c}{33.17\\0.882}&\tabincell{c}{32.29\\0.789}&\tabincell{c}{32.90\\0.845}&\tabincell{c}{27.74\\0.595}&\tabincell{c}{39.44\\0.960}&\tabincell{c}{29.29\\0.871}&\tabincell{c}{33.60\\0.812}&\tabincell{c}{31.53\\0.867}&\tabincell{c}{34.00\\0.811}&\tabincell{c}{28.60\\0.733}&\tabincell{c}
{32.26\\0.817}   \\
\midrule
BIRNAT \cite{BIRNAT_TPAMI2023} & TPMAI 2023 & 4.19 & 1973.66 &\tabincell{c}
{35.58\\0.914}&\tabincell{c}{35.23\\0.884}&\tabincell{c}{34.93\\0.864}&\tabincell{c}{30.23\\0.721}&\tabincell{c}{42.38\\0.976}&\tabincell{c}{35.97\\0.917}&\tabincell{c}{36.28\\0.896}&\tabincell{c}{33.62\\0.907}&\tabincell{c}{36.87\\0.888}&\tabincell{c}{30.48\\0.807}&\tabincell{c}{35.16\\0.877}  \\

\midrule
RDLUF \cite{Dong_2023_CVPR_rdluf} & CVPR 2023 & 1.81 & 115.34 &\tabincell{c}{37.06\\0.929}&\tabincell{c}{34.36\\0.863}&\tabincell{c}{\textbf{36.49}\\\textbf{0.891}}&\tabincell{c}{30.04\\0.687}&\tabincell{c}{\textbf{44.21}\\\textbf{0.983}}&\tabincell{c}{36.13\\0.929}&\tabincell{c}{36.78\\0.885}&\tabincell{c}{33.27\\0.903}&\tabincell{c}{37.04\\0.885}&\tabincell{c}
{32.02\\0.826}&\tabincell{c}{35.74\\0.878}  \\
\midrule
SSR-L \cite{Zhang_2024_CVPR_ssr} & CVPR 2024 & 5.18 & 78.93 &\tabincell{c}
{36.25\\0.921}&\tabincell{c}{33.60\\0.838}&\tabincell{c}{35.38\\0.880}&\tabincell{c}{29.01\\0.666}&\tabincell{c}{43.23\\0.978}&\tabincell{c}{36.41\\0.924}&\tabincell{c}{35.79\\0.863}&\tabincell{c}{33.57\\0.890}&\tabincell{c}{35.87\\0.860}&\tabincell{c}
{30.03\\0.745}&\tabincell{c}{34.91\\0.857} \\
\midrule 
DAUHST-SAH \cite{liu2024sahsci} & ECCV 2024 & 3.73 & 49.79 &\tabincell{c}
{35.10\\0.912}&\tabincell{c}{33.09\\0.812}&\tabincell{c}{34.84\\0.870}&\tabincell{c}{28.18\\0.593}&\tabincell{c}{42.82\\0.975}&\tabincell{c}{34.30\\0.918}&\tabincell{c}{34.60\\0.850}&\tabincell{c}{32.09\\0.867}&\tabincell{c}{34.54\\0.823}&\tabincell{c}
{29.86\\0.736}&\tabincell{c}{33.94\\0.836} \\
\midrule 
DPU-SAH  \cite{liu2024sahsci} & ECCV 2024 & 3.12 & 54.44 &\tabincell{c}
{36.93\\\textbf{0.932}}&\tabincell{c}{34.41\\0.853}&\tabincell{c}{36.04\\0.884}&\tabincell{c}{30.19\\0.697}&\tabincell{c}{43.43\\0.980}&\tabincell{c}{35.51\\0.930}&\tabincell{c}{36.20\\0.867}&\tabincell{c}{33.61\\0.895}&\tabincell{c}{36.80\\0.874}&\tabincell{c}
{31.02\\0.779}&\tabincell{c}{35.41\\0.869}   \\
\midrule
PSR-SCI \cite{zeng2025spectral} & ICLR 2025 & 1312 & 185702 &\tabincell{c}{34.48\\0.913}&\tabincell{c}{31.15\\0.765}&\tabincell{c}{35.58\\0.928}&\tabincell{c}{25.99\\0.527}&\tabincell{c}{36.20\\0.816}&\tabincell{c}{33.22\\0.920}&\tabincell{c}{32.12\\0.856}&\tabincell{c}{29.49\\0.839}&\tabincell{c}{32.14\\0.724}&\tabincell{c}{28.27\\0.650}&\tabincell{c}{31.86\\0.794} \\

\midrule \rowcolor{rouse}
\tabincell{c}{DAUHST-9stg\\(slow learning)} \cite{dauhst} & NeurIPS 2022 & 6.15 & 79.50 &\tabincell{c}
{34.95\\0.914}&\tabincell{c}{31.14\\0.758}&\tabincell{c}{34.67\\0.873}&\tabincell{c}{26.34\\0.532}&\tabincell{c}{42.56\\0.978}&\tabincell{c}{33.62\\0.920}&\tabincell{c}{32.64\\0.798}&\tabincell{c}{30.14\\0.845}&\tabincell{c}{32.51\\0.770}&\tabincell{c}
{27.81\\0.694}&\tabincell{c}{32.64\\0.808}  \\
\midrule 
\rowcolor{rouse}
\textbf{SlowFast-SCI$^1$}& \textbf{Ours} & 1.41 & 21.84 &\tabincell{c}
{35.52\\0.917}&\tabincell{c}{36.23\\0.868}&\tabincell{c}{33.91\\0.850}&\tabincell{c}{31.06\\0.694}&\tabincell{c}{42.25\\0.971}&\tabincell{c}{34.18\\0.910}&\tabincell{c}{37.89\\0.871}&\tabincell{c}{36.05\\0.909}&\tabincell{c}{37.80\\0.863}&\tabincell{c}
{33.16\\0.813}&\tabincell{c}{35.81\\0.867}  \\ 

\midrule \rowcolor{rouse}
\tabincell{c}{DPU-9stg\\(slow learning)} \cite{Zhang_2024_CVPR_dpu} & CVPR 2024 & 2.85 & 49.26 &\tabincell{c}{36.90\\\textbf{0.932}}&\tabincell{c}{34.36\\0.852}&\tabincell{c}{36.02\\0.884}&\tabincell{c}{30.14\\0.695}&\tabincell{c}{43.41\\0.980}&\tabincell{c}{35.45\\0.930}&\tabincell{c}{36.14\\0.866}&\tabincell{c}{33.56\\0.894}&\tabincell{c}{36.72\\0.873}&\tabincell{c}
{30.97\\0.777}&\tabincell{c}{35.37\\0.868}  \\
\midrule
\rowcolor{rouse}
\textbf{SlowFast-SCI$^2$}& \textbf{Ours} & 0.70 & 15.17 
&\tabincell{c}{\textbf{37.15}\\0.930}&\tabincell{c}{\textbf{41.93}\\\textbf{0.949}}&\tabincell{c}{36.15\\0.879}
&\tabincell{c}{\textbf{37.94}\\\textbf{0.865}}&\tabincell{c}{44.09\\0.978}&\tabincell{c}{\textbf{37.65}\\\textbf{0.934}}
&\tabincell{c}{\textbf{45.88}\\\textbf{0.971}}&\tabincell{c}{\textbf{42.25}\\\textbf{0.965}}&\tabincell{c}{\textbf{45.52}\\\textbf{0.960}}
&\tabincell{c}{\textbf{43.09}\\\textbf{0.962}}&\tabincell{c}{\textbf{41.16}\\\textbf{0.939}}\\
\bottomrule[0.1em]
\end{tabular}
}
\label{tab:harvard}
\end{table*}

\begin{table*}[t]
\def\arraystretch{1.1}
\setlength{\tabcolsep}{5pt}
\newcommand{\tabincell}[2]
{\begin{tabular}{@{}#1@{}}#2\end{tabular}}
\centering
\caption{Comparisons (PSNR (upper entry ineach cell), SSIM (lower entry in each cell), Params, and FLOPs) between SlowFast-SCI and SOTA methods on 10 simulation scenes of \texttt{ICVL} dataset.
}
\resizebox{0.99\textwidth}{!}
{
\centering
\begin{tabular}{ccccccccccccccc}
\toprule[0.1em]
\rowcolor{lightgray}
Method & Reference & Params (M) & FLOPs (G) & S1 & S2 & S3 & S4 & S5 & S6 & S7 & S8 & S9 & S10 & Avg \\
\midrule
TSA-Net \cite{meng2020endtsa_cassi} & ECCV 2020 & 42.20 & 91.58 &\tabincell{c}{26.22\\0.720}&\tabincell{c}{28.50\\0.736}&\tabincell{c}{28.02\\0.777}&\tabincell{c}{26.32\\0.819}&\tabincell{c}{31.10\\0.736}&\tabincell{c}{28.23\\0.750}&\tabincell{c}{20.91\\0.783}&\tabincell{c}{25.14\\0.762}&\tabincell{c}{28.74\\0.700}&\tabincell{c}
{27.25\\0.817}&\tabincell{c}{27.05\\0.760}   \\

\midrule
MST \cite{mst} & CVPR 2022 & 1.92 & 25.60 &\tabincell{c}{27.07\\0.877}&\tabincell{c}{31.89\\0.909}&\tabincell{c}{31.80\\0.883}&\tabincell{c}{30.43\\0.900}&\tabincell{c}{35.79\\0.931}&\tabincell{c}{32.22\\0.893}&\tabincell{c}{24.05\\0.829}&\tabincell{c}{27.00\\0.842}&\tabincell{c}{34.93\\0.907}&\tabincell{c}{26.30\\0.815}&\tabincell{c}
{30.15\\0.879}   \\
\midrule
CST \cite{cst} & ECCV 2022 & 2.86 & 36.77 &\tabincell{c}{26.61\\0.871}&\tabincell{c}{31.42\\0.904}&\tabincell{c}{31.97\\0.885}&\tabincell{c}{29.36\\0.881}&\tabincell{c}{36.14\\0.935}&\tabincell{c}{32.00\\0.884}&\tabincell{c}{23.26\\0.792}&\tabincell{c}{26.35\\0.833}&\tabincell{c}{35.24\\0.910}&\tabincell{c}{26.72\\0.846}&\tabincell{c}
{29.91\\0.874}  \\
\midrule
BIRNAT \cite{bisci} & TPMAI 2023 & 4.19 & 1973.66 &\tabincell{c}
{34.78\\0.934}&\tabincell{c}{38.41\\0.955}&\tabincell{c}{36.93\\0.951}&\tabincell{c}{38.52\\0.957}&\tabincell{c}{42.17\\0.970}&\tabincell{c}{37.40\\0.945}&\tabincell{c}{30.35\\0.926}&\tabincell{c}{32.59\\0.906}&\tabincell{c}{37.73\\0.932}&\tabincell{c}{34.87\\0.938}&\tabincell{c}{36.37\\0.941}  \\

\midrule
RDLUF \cite{Dong_2023_CVPR_rdluf} & CVPR 2023 & 1.81 & 115.34 &\tabincell{c}{36.67\\0.953}&\tabincell{c}{41.16\\\textbf{0.970}}&\tabincell{c}{40.79\\0.975}&\tabincell{c}{40.93\\0.971}&\tabincell{c}{\textbf{45.72}\\\textbf{0.983}}&\tabincell{c}{39.34\\0.959}&\tabincell{c}{34.12\\0.956}&\tabincell{c}{34.73\\0.928}&\tabincell{c}{\textbf{39.87}\\\textbf{0.951}}&\tabincell{c}
{37.27\\0.960}&\tabincell{c}{39.06\\0.961} \\
\midrule
SSR-L \cite{Zhang_2024_CVPR_ssr} & CVPR 2024 & 5.18 & 78.93  &\tabincell{c}{35.77\\0.938}&\tabincell{c}{41.14\\0.968}&\tabincell{c}{39.39\\0.966}&\tabincell{c}{41.20\\0.970}&\tabincell{c}{45.53\\0.982}&\tabincell{c}{\textbf{39.69}\\\textbf{0.960}}&\tabincell{c}{33.24\\0.947}&\tabincell{c}{34.72\\0.922}&\tabincell{c}{38.90\\0.942}&\tabincell{c}
{37.30\\0.946}&\tabincell{c}{38.69\\0.954} \\
\midrule 
DAUHST-SAH \cite{liu2024sahsci} & ECCV 2024 & 3.73 & 49.79 &\tabincell{c}
{29.94\\0.902}&\tabincell{c}{34.24\\0.937}&\tabincell{c}{34.50\\0.936}&\tabincell{c}{32.88\\0.934}&\tabincell{c}{39.74\\0.962}&\tabincell{c}{34.98\\0.929}&\tabincell{c}{26.37\\0.893}&\tabincell{c}{29.82\\0.879}&\tabincell{c}{37.23\\0.926}&\tabincell{c}
{28.39\\0.878}&\tabincell{c}{32.81\\0.918} \\
\midrule 
DPU-SAH \cite{liu2024sahsci} & ECCV 2024 & 3.12 & 54.44 &\tabincell{c}
{33.90\\0.940}&\tabincell{c}{39.55\\0.966}&\tabincell{c}{38.12\\0.971}&\tabincell{c}{39.82\\0.967}&\tabincell{c}{45.26\\0.981}&\tabincell{c}{39.36\\0.960}&\tabincell{c}{28.31\\0.919}&\tabincell{c}{32.63\\0.918}&\tabincell{c}{39.76\\0.951}&\tabincell{c}{35.85\\0.950}&\tabincell{c}{37.26\\0.952}  \\
\midrule
PSR-SCI \cite{zeng2025spectral} & ICLR 2025 & 1312 & 185702 &\tabincell{c}{30.92\\0.881}&\tabincell{c}{36.71\\0.955}&\tabincell{c}{35.10\\0.929}&\tabincell{c}{36.77\\0.970}&\tabincell{c}{43.27\\0.978}&\tabincell{c}{33.93\\0.911}&\tabincell{c}{27.27\\0.876}&\tabincell{c}{32.86\\0.926}&\tabincell{c}{36.14\\0.925}&\tabincell{c}{33.25\\0.940}&\tabincell{c}{34.62\\0.929} \\

\midrule \rowcolor{rouse}
\tabincell{c}{DAUHST-9stg\\(slow learning)} \cite{dauhst} & NeurIPS 2022 & 6.15 & 79.50 &\tabincell{c}
{29.49\\0.902}&\tabincell{c}{34.26\\0.941}&\tabincell{c}{34.43\\0.938}&\tabincell{c}{32.83\\0.938}&\tabincell{c}{39.94\\0.966}&\tabincell{c}{35.07\\0.932}&\tabincell{c}{25.59\\0.890}&\tabincell{c}{29.61\\0.883}&\tabincell{c}{37.45\\0.932}&\tabincell{c}{28.59\\0.883}&\tabincell{c}{32.73\\0.920} \\
\midrule 
\rowcolor{rouse}
\textbf{SlowFast-SCI$^1$}& \textbf{Ours} & 1.41 & 21.84  &\tabincell{c}
{35.48\\0.936}&\tabincell{c}{39.99\\0.964}&\tabincell{c}{38.91\\0.963}&\tabincell{c}{39.73\\0.964}&\tabincell{c}{43.86\\0.975}&\tabincell{c}{37.55\\0.946}&\tabincell{c}{32.23\\0.938}&\tabincell{c}{34.31\\0.921}&\tabincell{c}{38.40\\0.934}&\tabincell{c}
{36.35\\0.952}&\tabincell{c}{37.68\\0.950}  \\ 

\midrule \rowcolor{rouse}
\tabincell{c}{DPU-9stg\\(slow learning)} \cite{Zhang_2024_CVPR_dpu} & CVPR 2024 & 2.85 & 49.26 &\tabincell{c}{33.88\\0.939}&\tabincell{c}{39.51\\0.966}&\tabincell{c}{38.09\\0.971}&\tabincell{c}{39.76\\0.966}&\tabincell{c}{45.21\\0.981}&\tabincell{c}{39.32\\\textbf{0.960}}&\tabincell{c}{28.29\\0.919}&\tabincell{c}{32.61\\0.918}&\tabincell{c}{39.74\\\textbf{0.951}}&\tabincell{c}{35.81\\0.950}&\tabincell{c}{37.22\\0.952}  \\
\midrule
\rowcolor{rouse}
\textbf{SlowFast-SCI$^2$}& \textbf{Ours} & 0.70 & 15.17 
&\tabincell{c}{\textbf{38.59}\\\textbf{0.962}}&\tabincell{c}{\textbf{41.33}\\\textbf{0.970}}&\tabincell{c}{\textbf{41.54}\\\textbf{0.979}}
&\tabincell{c}{\textbf{42.15}\\\textbf{0.972}}&\tabincell{c}{45.49\\0.981}&\tabincell{c}{39.22\\0.957}
&\tabincell{c}{\textbf{35.72}\\\textbf{0.957}}&\tabincell{c}{\textbf{36.16}\\\textbf{0.937}}&\tabincell{c}{39.45\\0.949}
&\tabincell{c}{\textbf{38.26}\\\textbf{0.962}}&\tabincell{c}{\textbf{39.79}\\\textbf{0.963}} \\ 
\bottomrule[0.1em]
\end{tabular}
}
\label{tab:icvl}
\end{table*}

\textbf{Formal definitions and a schematic illustration of the four loss terms.}
The introduced self-supervised training loss is motivated by \cite{liu2024sahsci}
\begin{equation}
    \mathscr{L}_{sst} = \mathscr{L}_{m} + \omega_1\mathscr{L}_{ei} + \omega_2\mathscr{L}_{iu} + \omega_3\mathscr{L}_{tv},
\end{equation}
where $\mathscr{L}_m$, $\mathscr{L}_{ei}$, $\mathscr{L}_{iu}$, and $\mathscr{L}_{tv}$ denote the measurement consistency loss, the equivariance imaging loss, the image uncertainty loss, 
and the total variation loss \cite{rudin1992nonlinear_tv}, respectively, as illustrated in Fig. \ref{fig:loss}.

Since the measurement $\mathrm{\mathbf{y}}_i$ is encoded from the hyperspectral image $\mathrm{\mathbf{x}}_i$ via 
$\mathbf{\Phi}\mathrm{\mathbf{x}}_i$, we treat 
$\mathrm{\mathbf{y}}_i$ as the label for self-supervised training. We first enforce that the inverse mapping 
$\mathcal{F}_{AdaptUN}(\cdot;\omega^*, \theta)$ remains consistent in the measurement domain.
Thus the measurement consistency loss $\mathscr{L}_{m}$ is defined as follows:
\begin{equation}
    \mathscr{L}_m(\theta) = \frac{1}{N_d} \sum_{i=1}^{N_d} \| \mathrm{\mathbf{y}}_i - \mathbf{\Phi}\mathcal{F}_{AdaptUN}(\mathrm{\mathbf{y}}_i;\omega^*, \theta) \|_2^2.
\end{equation}
The $\mathscr{L}_{ei}$ denotes the equivariance imaging loss:
\begin{equation}
\begin{split}
    \mathscr{L}_{ei}(\theta) &= \frac{1}{|\mathcal{G}| N_d} \sum_{i=1}^{N_d} \sum_{g=1}^{|\mathcal{G}|} 
    \| T_g \mathcal{F}_{AdaptUN}(\mathrm{\mathbf{y}}_i;\omega^*, \theta) \\
    & - \mathcal{F}_{AdaptUN}(\mathbf{\Phi} T_g \mathcal{F}_{AdaptUN}(\mathrm{\mathbf{y}}_i;\omega^*, \theta);\omega^*, \theta) \|_2^2,
\end{split}
\end{equation}
where $\left\{ T_g \right\}_{g=1,...,|\mathcal{G}|}$ denotes a group of transformations (\textit{e.g.}, shifts, rotations, \textit{etc.}), which are introduced to mitigate the limitations caused by a single incomplete measurement mask in the SCI system.

The null space of the operator leads to infinitely many feasible solutions, making the reconstructed hyperspectral images ambiguous. The image uncertainty loss $\mathscr{L}{iu}$ addresses this ambiguity by modeling reconstruction uncertainty \cite{quan2022dual}:
\begin{equation}
    \mathscr{L}_{iu}(\theta) = \frac{1}{N_d} \sum_{i=1}^{N_d} \| \mathcal{F}_{AdaptUN}(\mathbf{\Phi}(\mathrm{\mathbf{z}}_i + \mathrm{\mathbf{e}});\omega^*, \theta) - \mathrm{\mathbf{z}}_i + \mathrm{\mathbf{e}} \|_2^2,
\end{equation}
where $\mathrm{\mathbf{z}}_i=\mathcal{F}_{AdaptUN}(\mathrm{\mathbf{y}}_i;\omega^*, \theta)$ and $\mathrm{\mathbf{e}}$ is generated by random sign-flipping based on the approximate calculation of the real residual.

The total variation loss $\mathscr{L}_{tv}$ is introduced to remove noise and eliminate artifacts that may arise during HSI reconstruction: 
\begin{equation}
    \mathscr{L}_{tv}(\theta) = \frac{1}{N_d} \sum_{i=1}^{N_d} TV(\mathcal{F}_{AdaptUN}(\mathrm{\mathbf{y}}_i;\omega^*, \theta)). 
\end{equation}
Let $\hat{\theta}$ denote the parameters of the adapt layers after self-supervised fine-tuning.
Then, given a test sample $\mathrm{\mathbf{\tilde{y}}} \in \{\mathrm{\mathbf{y}}_i\}_{i=1}^{N_t}$, 
we adapt the model $\mathcal{F}_{AdaptUN}(\cdot;\omega^*, \hat{\theta})$ to $\mathrm{\mathbf{\tilde{y}}}$ by minimizing self-supervised loss $\mathscr{L}_{tta}$:
\begin{equation}
    \mathscr{L}_{tta} = \mathscr{L}_{m} + \alpha\mathscr{L}_{ei} + \beta\mathscr{L}_{tv},
    \label{l_tta}
\end{equation}
where $\alpha$ and $\beta$ denote the equilibrium constant. 
Both $\mathscr{L}_{ei}$ (equivariance imaging loss) and $\mathscr{L}_{iu}$ (image uncertainty loss) successfully overcome the reconstruction ambiguity stemming from the infinite solutions of the uninformative nullspace of the operator \cite{liu2024sahsci}. Nevertheless, because the TTA stage demands rapid updating, we exclusively utilize the computationally lighter $\mathscr{L}_{ei}$.

\begin{algorithm}
\caption{Inference with Test-Time Adaptation}
\textbf{Input:} Measurement sample $\mathrm{\mathbf{\tilde{y}}}$; Sensing matrix $\mathbf{\Phi}$; Pre-trained backbone parameters $\boldsymbol{\omega}^*$; Self-supervised pre-trained test-time adaptation module parameters $\boldsymbol{\theta}^*$ \\
$\mathcal{F}_{AdaptUN}(\cdot;\boldsymbol{\omega}^*, \boldsymbol{\theta}^*)$ represents the model; $T(\cdot)$ represents the transformation \\
\textbf{Parameter:} Learning rate $\tau$; Trade-off hyperparameter $\alpha$ and $\beta$; Iteration number $K$ \\
\textbf{Output:} Reconstructed image $\mathrm{\mathbf{\tilde{x}}}$
\begin{algorithmic}[1]
\State Initialize $\boldsymbol{\tilde{\theta}}^*$ with $\boldsymbol{\theta}^*$.
\For{$i = 1, \cdots, K$}, update $\boldsymbol{\tilde{\theta}}^*$ on test sample:
    \State $\mathrm{\mathbf{\tilde{x}}}_i = \mathcal{F}_{AdaptUN}(\mathrm{\mathbf{\tilde{y}}};\boldsymbol{\omega}^*, \boldsymbol{\tilde{\theta}}^*)$
    \State $\mathscr{L}_{m}(\mathrm{\mathbf{\tilde{y}}};\boldsymbol{\omega}^*, \boldsymbol{\tilde{\theta}}^*) = \| \mathrm{\mathbf{\tilde{y}}} - \mathbf{\Phi}\mathrm{\mathbf{\tilde{x}}}_i \|_2^2$ 
    \State $\mathscr{L}_{ei}(\mathrm{\mathbf{\tilde{y}}};\boldsymbol{\omega}^*, \boldsymbol{\tilde{\theta}}^*) =  
    \| T (\mathrm{\mathbf{\tilde{x}}}_i) - \mathcal{F}_{AdaptUN}(\mathbf{\Phi} T (\mathrm{\mathbf{\tilde{x}}}_i); \boldsymbol{\omega}^*, \boldsymbol{\tilde{\theta}}^*) \|_2^2$
    \State $\mathscr{L}_{tv}(\mathrm{\mathbf{\tilde{y}}};\boldsymbol{\omega}^*, \boldsymbol{\tilde{\theta}}^*) = TV(\mathrm{\mathbf{\tilde{x}}}_i)$ 
    \State $\boldsymbol{\tilde{\theta}}^* := \boldsymbol{\tilde{\theta}}^* - \tau \nabla_{\boldsymbol{\tilde{\theta}}^*} \mathscr{L}_{m}(\mathrm{\mathbf{\tilde{y}}};\boldsymbol{\omega}^*, \boldsymbol{\tilde{\theta}}^*) - \tau\alpha\nabla_{\boldsymbol{\tilde{\theta}}^*}\mathscr{L}_{ei}(\mathrm{\mathbf{\tilde{y}}};\boldsymbol{\omega}^*, \boldsymbol{\tilde{\theta}}^*) -
    \tau\beta\nabla_{\boldsymbol{\tilde{\theta}}^*}\mathscr{L}_{tv}(\mathrm{\mathbf{\tilde{y}}};\boldsymbol{\omega}^*, \boldsymbol{\tilde{\theta}}^*)$
\EndFor
\State \textbf{return} $\mathrm{\mathbf{\tilde{x}}} = \frac{1}{K} 
\sum_{i=1}^K \mathrm{\mathbf{\tilde{x}}}_i$
\end{algorithmic}
\end{algorithm}

\begin{figure*}[h]
    \centering
    \includegraphics[width=0.98\linewidth]{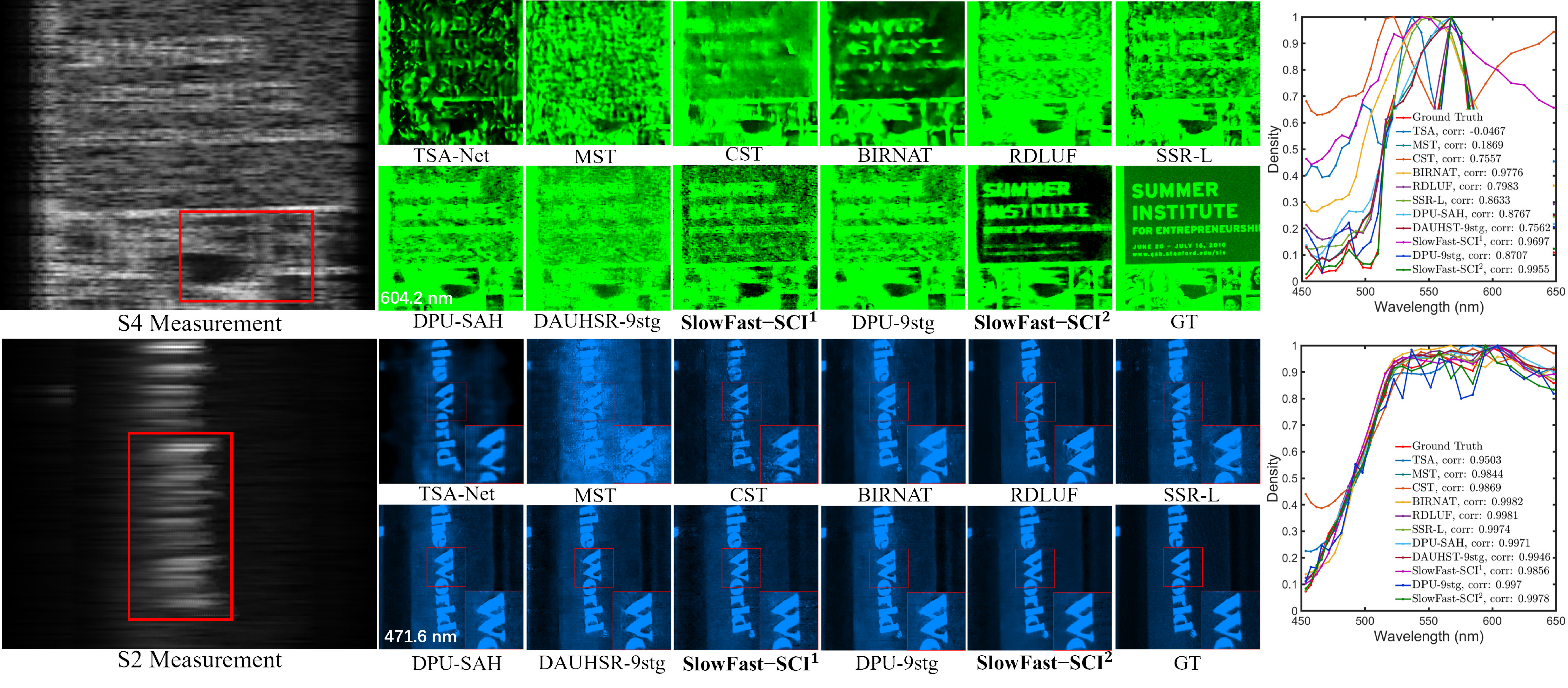}
    \vspace{-4mm}
    \caption{Simulated HSI reconstruction comparisons (middle) on \texttt{Harvard}. The left shows the scene measurement and the right shows the spectral curves corresponding to the selected region. Please zoom in for better view.
    }
    \label{fig:harvard}
    \vspace{-4mm}
\end{figure*}
\begin{figure*}[h]
    \centering
    \includegraphics[width=0.98\linewidth]{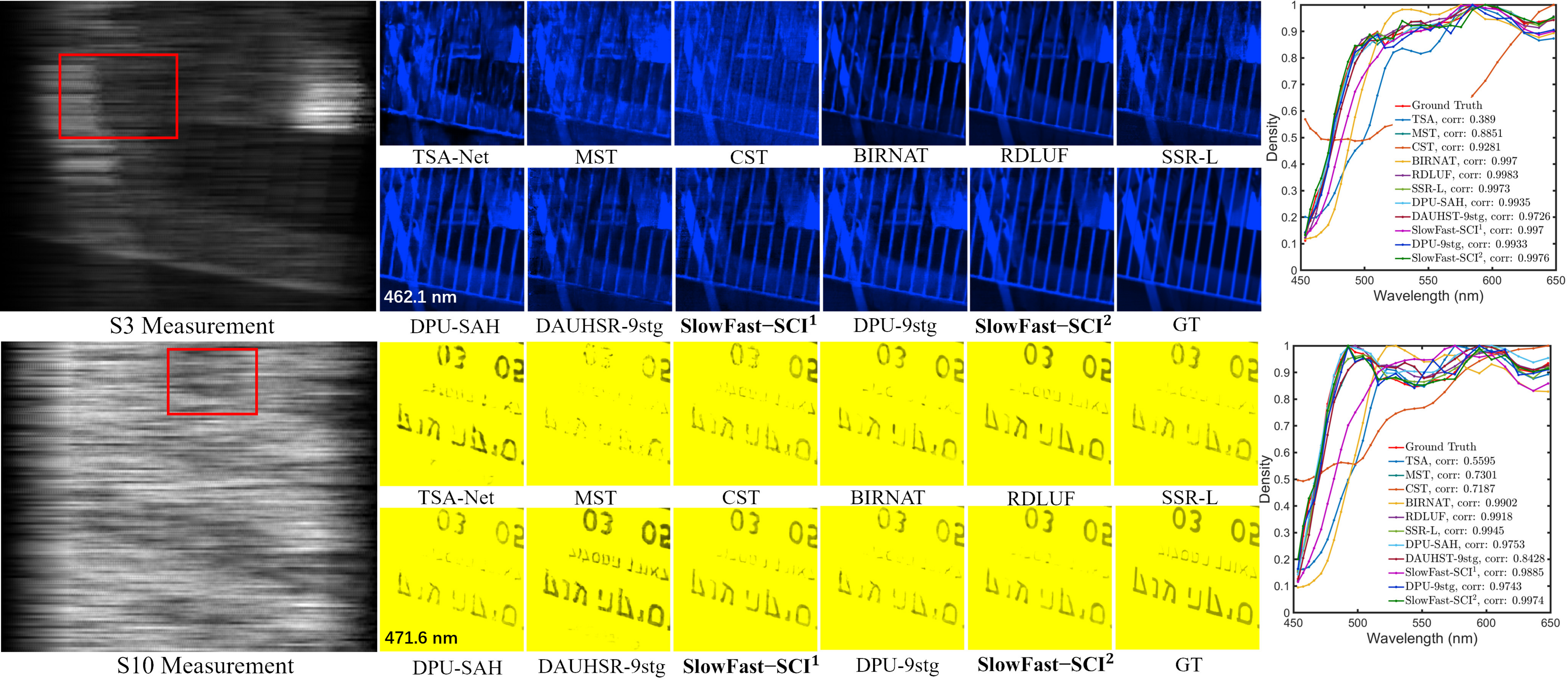}
    \vspace{-4mm}
    \caption{
    Simulated HSI reconstruction comparisons (middle) on \texttt{ICVL}. The left shows the scene measurement and the right shows the spectral curves corresponding to the selected region. 
    Please zoom in for better view.
    }
    \label{fig:icvl}
    \vspace{-6mm}
\end{figure*}

\begin{figure*}[t]
    \centering
    \includegraphics[width=0.98\linewidth]{Fig/real.pdf}
    \caption{Reconstructed images of real scene 1 and 2 with 2 out of 28 spectral channels by the state-of-the-art methods. Compared with other
    methods, our SlowFast-SCI recovers more details and clear content.
    Please zoom in for better view.}
    \label{fig:real}
\end{figure*}

\vspace{-2mm}

\section{Experiments}
\subsection{Experiment Setup and Implementation Details}

\textbf{Datasets.}
The pre-trained model used in slow learning was trained on the \texttt{CAVE} \cite{park2007multispectralcave} dataset.
We adopted two datasets \texttt{Harvard} \cite{chakrabarti2011statistics_harvard} and \texttt{ICVL} \cite{arad2016sparse_icvl} with randomly clipped
patches of size $256\times256\times28$ 
for simulation experiments. 
\texttt{Harvard/ICVL} dataset contains 67/180 images for training and 10/20 for testing.
For real datasets, we used 5 scenes with spatial size $660\times660$ and 28 bands captured in TSA-Net \cite{meng2020endtsa_cassi}, then cropped data blocks of size $256\times256\times28$.
\textbf{Comparison Methods.} We compare our SlowFast-SCI on synthetic data and real scenes with state-of-the-art (SOTA) reconstruction methods including TSA-Net \cite{meng2020endtsa_cassi}, MST\cite{mst},  CST\cite{cst}, BIRNAT\cite{bisci}, RDLUF\cite{Dong_2023_CVPR_rdluf}, SSR-L \cite{Zhang_2024_CVPR_ssr}, SAH-SCI \cite{liu2024sahsci}, PSR-SCI \cite{zeng2025spectral},
DAUHST \cite{dauhst}, and DPU \cite{Zhang_2024_CVPR_dpu}.

\textbf{Image-guided fast unfolding distillation and self-supervised fine-tuning.} 
In IGFUD, we distilled two typical DUNs with 9 stages, DAUHST \cite{darestani2022test_tta} and DPU \cite{Zhang_2024_CVPR_dpu}.
The distillation epoch and batch size were set to 200 and 2, respectively.
We used the Adam \cite{2014Adam} optimizer for both networks.
We set the initial learning rate as $4\times{10}^{-4}$ and adopted the Cosine
Annealing learning rate scheme \cite{loshchilov2016sgdr} to implement  training.
We fine-turned the FAST-TTAM for 100 epochs. The batch size was set to 2 and the learning rate was fixed at $1\times{10}^{-4}$. The $\omega_1$, $\omega_2$ and $\omega_3$ were set to 0.7, 0.4 and 0.001 in Eq. (\ref{l_self}).

\textbf{Test-Time Adaptation.} The test-time adaptation was conducted over 50 iterations for \texttt{Harvard} and 10
iterations for \texttt{ICVL}, with a constant learning rate of $1\times{10}^{-3}$ and $\alpha = 0.7, \beta=0.001$ throughout all iterations.


\subsection{Quantitative Results} 


\paragraph{Comparison on \texttt{Harvard} and \texttt{ICVL}.}
Tabs.~\ref{tab:harvard} and~\ref{tab:icvl} summarize PSNR, SSIM, Params, and FLOPs for SlowFast-SCI$^1$ (slow-learning backbone: DAUHST) and SlowFast-SCI$^2$ (slow-learning backbone: DPU) against SOTA methods.  
SlowFast-SCI$^2$ achieves 41.16dB/0.939 on \texttt{Harvard} and 39.79dB/0.963 on \texttt{ICVL}, outperforming all competitors.  
SlowFast-SCI$^1$ yields 35.81dB/0.867 and 37.68dB/0.950, improving its backbone by 9.71\%/7.30\% on \texttt{Harvard} and 15.12\%/3.26\% on \texttt{ICVL}. 
Both variants use $<$25\% of the Params and $<$33\% of the FLOPs of their backbones.  
Compared to SAH-SCI, SlowFast-SCI$^2$ achieves 16.24\%/8.06\% and 6.79\%/1.16\% gains on \texttt{Harvard} and \texttt{ICVL}, respectively, with lower resource costs; SlowFast-SCI$^1$ attains similar improvements.

\subsection{Qualitative Results}
\textbf{Simulation Data Results.}
Figs. \ref{fig:harvard} and \ref{fig:icvl} depict the simulation HSI reconstruction comparisons between our method and other SOTA methods on \texttt{Harvard} and \texttt{ICVL} datasets, respectively.
It can be seen that the other models produce
poor reconstruction results with a lot of noise and artifacts, since they lack fast learning on similar data.
The results of our SlowFast-SCI in different bands show that the proposed dual speed framework can correct the problem of weak generalization ability of SCI reconstruction due to insufficient training data.
This is because our method enhances the generalization of the model to extract more spectral information through self-supervised fast learning and can adapt the model to align it with target scenes at inference. 
Besides, the spectral density curves show that the reconstruction results of SlowFast-SCI are more similar and correlated with the GT, which proves our method is capable of improving the spectral-dimension consistency of the model.

\begin{figure}[t]
    \centering
    \includegraphics[width=0.98\linewidth]{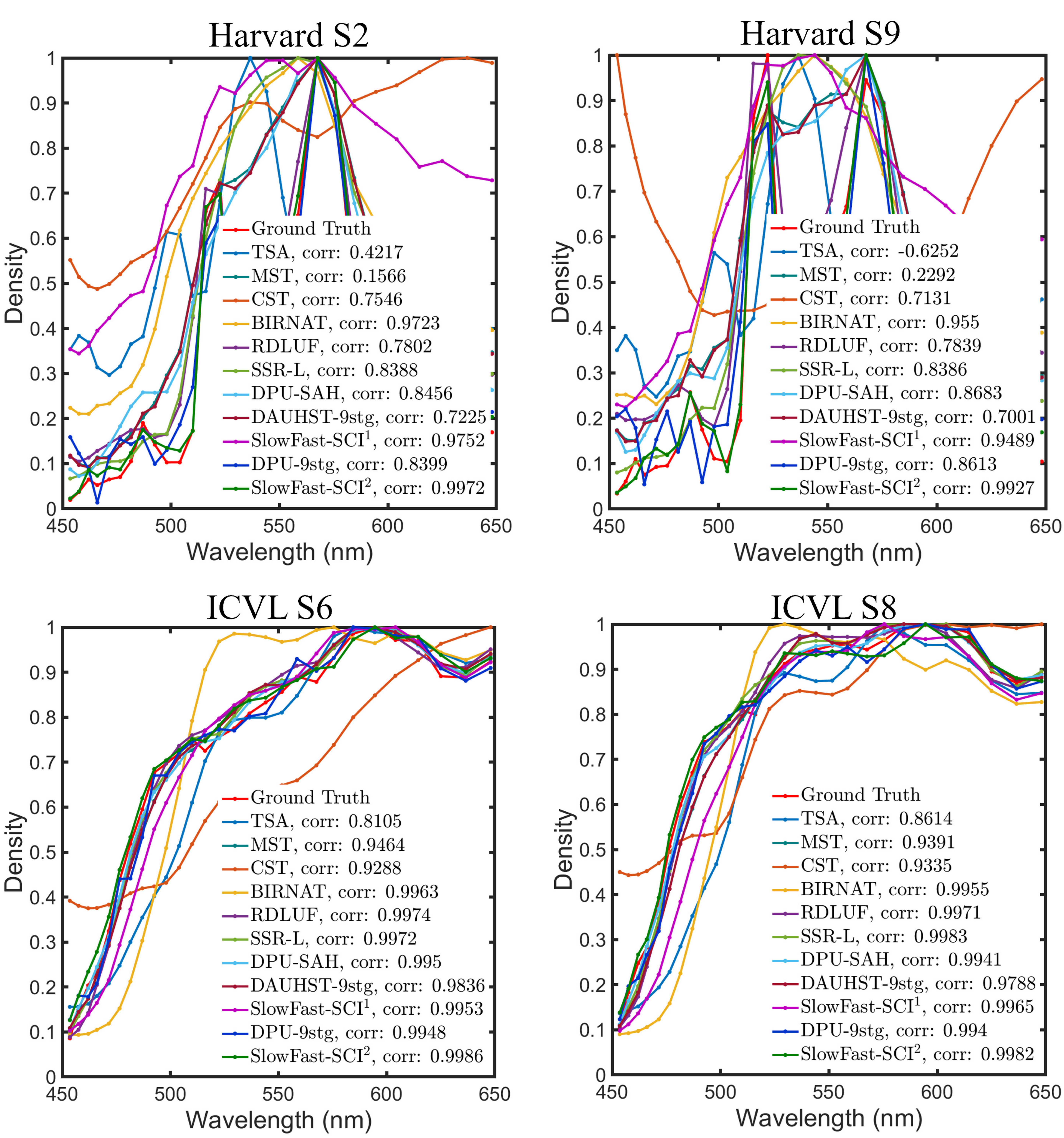}
    \caption{The spectral curves corresponding to the selected region in Figs.~\ref{fig:harvard1}, \ref{fig:harvard2}, \ref{fig:icvl1} and \ref{fig:icvl2}. The subplots, arranged in a $2\times2$ grid from the upper left to the lower right, correspond to \texttt{Harvard} Scenes 2 and 9, and \texttt{ICVL} Scenes 6 and 8, respectively.
    }
    \label{fig:spe_curve}
\end{figure}


\textbf{Real-Scene HSI Reconstruction.}
We evaluate SlowFast-SCI on real-world HSI using \texttt{CAVE} and \texttt{KAIST} \cite{choi2017highkaist}, pre-training a two-stage DPU backbone with real CASSI masks under the protocols of Zhang et al.~\cite{Zhang_2024_CVPR_dpu}, Meng et al.~\cite{meng2020endtsa_cassi}, and Cai et al.~\cite{dauhst}. To simulate realistic noise, we inject 11-bit shot noise during both slow- and fast-learning. At test time, raw measurements serve as self-supervised labels for adaptation. As shown in Fig.~\ref{fig:real}, SlowFast-SCI recovers finer details and suppresses noise more effectively than SOTAs, demonstrating strong real-world potential.

\begin{table}[t]
\centering
\caption{Ablation on the stages of fast unfolding network,
FAST-TTAM, and test-time adaptation.}
\label{tab:ablations1}
\def\arraystretch{1.1}
\setlength{\tabcolsep}{2.4pt}
\resizebox{\linewidth}{!}{%
\begin{tabular}{c c c c | c c | c c | c}
\toprule
Case & Stages & FAST-TTAM & TTA & PSNR & SSIM
& Params (M) & FLOPs (G) & Time (s) \\
\midrule
(a) & 9 & $\times$   & $\times$   & 35.37 & 0.868 & 2.85 & 49.26 & $\sim$0.22 \\
(b) & 9 & \checkmark & $\times$   & 34.93 & 0.850 & 3.16 & 67.92 & $\sim$0.25 \\
(c) & 9 & \checkmark & \checkmark & 35.00 & 0.851 & 3.16 & 67.92 & $\sim$7.93 \\
(d) & 1 & $\times$   & $\times$   & 35.67 & 0.876 & 0.32 & 5.39  & $\sim$0.10 \\
(e) & 1 & \checkmark & $\times$   & 38.39 & 0.912 & 0.35 & 7.63  & $\sim$0.10 \\
(f) & 1 & \checkmark & \checkmark & 38.39 & 0.903 & 0.35 & 7.63  & $\sim$1.61 \\
(g) & 2 & $\times$   & $\times$   & 35.75 & 0.876 & 0.64 & 10.86 & $\sim$0.12 \\
(h) & 2 & \checkmark & $\times$   & 39.97 & 0.932 & 0.70 & 15.17 & $\sim$0.12 \\
(i) & 2 & \checkmark & \checkmark & 40.33 & 0.931 & 0.70 & 15.17 & $\sim$1.82 \\
\bottomrule
\end{tabular}%
}
\end{table}

\begin{figure}[t]
\vspace{-2mm}
\centering
\includegraphics[width=\linewidth]{Fig/iter_new.pdf}
\vspace{-6mm}
\caption{Ablation on the number of inference iterations.}
\vspace{-2mm}
\label{fig:iter}
\end{figure}

\subsection{Ablation Studies}

\textbf{Iteration Times at Inference.} This experiment investigates the relationship
between iteration times and performance gains through TTA.
Fig. \ref{fig:iter} shows 
that as the number of iterations increases, the PSNR improves.
Only a few iterations ($\approx$1.8s) are needed to achieve a substantial improvement (+0.36dB).
This confirms that our adaptation mechanism is highly effective in early stages.

\textbf{Fast Unfolding-Stage Configuration Analysis.}
We assess the effect of different unfolding stages on performance using the \texttt{Harvard} dataset (Tab.~\ref{tab:ablations1}), where FAST-TTAM denotes the unfolding stage attached with the adaptation module, and TTA is performed with 10 iterations.
Using the 9-stage DPU as a baseline (case a), results from cases (a, d, g) show that both 1-stage and 2-stage variants preserve the key characteristics of the original model.
However, cases (e, h) and (f, i) show that the 1-stage variant learns less effectively than the 2-stage network. Thus, despite its faster inference, the 1-stage model is discarded in favor of the better-performing 2-stage variant that enables a balance between reconstruction quality and inference speed. 
In particular, case (c) represents the AdaptNet \cite{eccv2024_tta} baseline adapted to the SCI configuration,
yet it suffers from poor OOD generalization and high latency. 
Conversely, our dual-speed framework (case i) not only facilitates robust test-time alignment with OOD samples but also achieves a substantial speedup in inference efficiency.

\textbf{Fast Learning Ablation.} We analyze the performance of FAST-TTAM and TTA in the fast learning. Tab. \ref{tab:ablations1} (d, e) and (g, h) show that the proposed FAST-TTAM can effectively transfer the fast unfolding variant to the target domain through self-supervised fine-tuning. Besides, case (h, i) in Tab.
\ref{tab:ablations1} shows self-supervised TTA can further align the fast unfolding variant with target scenes. 

\textbf{Fast test-time adaptation module ablation.} We analyze the performance of different components of the lightweight adaptation module on \texttt{Harvard} dataset in Tab. \ref{tab:sup_ablations2}, where Spatial refers to the spatial convolution block and Spectral refers to the spectral convolution block. Case (a) uses the original two-stage fast unfolding network.
As presented in Tab.~\ref{tab:sup_ablations2}, incorporating the proposed spatial convolution block leads to a PSNR improvement of 3.97dB (\textit{cf.} cases (a) and (b)), while the spectral convolution block yields a gain of 2.44dB (\textit{cf.} cases (a) and (c)). The results of cases (b), (c), and (d) further substantiate the effectiveness of jointly employing spatial and spectral convolution modules.

 \begin{table}[t]
  \centering
  \caption{Ablation on different components of the fast test-time adaptation module.}
  \label{tab:sup_ablations2}
  \def\arraystretch{1.1}
  \setlength{\tabcolsep}{2.4pt}
  \resizebox{0.98\columnwidth}{!}
  {
  \begin{tabular}{c c c c c c c}
  \toprule
  Case & Spatial & Spectral & PSNR & SSIM & Params (M) & FLOPs (G) \\
  \midrule
  (a) & $\times$&$\times$& 35.75  & 0.876  & 0.6356  & 10.86  \\
  (b) & \checkmark&$\times$& 39.72  & 0.931  & 0.7030  & 15.07  \\
  (c) & $\times$&\checkmark&  38.19 & 0.917  &  0.6429  & 11.23 \\
  (d) & \checkmark&\checkmark& 39.97 & 0.932  & 0.7045  & 15.17  \\
  \bottomrule
  \end{tabular}
  }
\end{table}

\emph{We provide additional visual comparisons of SlowFast-SCI against SOTA methods on simulated and real
scenes (Figs.~\ref{fig:harvard1}, \ref{fig:harvard2}, \ref{fig:icvl1}, \ref{fig:icvl2} and \ref{fig:app_real}) and spectral-curve analyses (Fig. \ref{fig:spe_curve}).}

\section{Conclusion}

We presented SlowFast-SCI, to our knowledge the first test-time-adaptation--driven deep-unfolding framework for spectral compressive imaging. 
Inspired by the complementary slow and fast pathways of human learning, our design couples a knowledge-distilled, prior-aware compact backbone (slow stage) with stage-wise lightweight FAST-TTAMs that are fine-tuned self-supervised at inference (fast stage)---without retraining the backbone. 
It jointly tackles two persistent limitations of unfolding pipelines: their poor generalization to unseen optical configurations and the heavy computational cost of deep stage cascades.
SlowFast-SCI cuts parameters and FLOPs by over 70\%, attains up to 5.79\,dB PSNR gain, preserves strong cross-domain adaptability, and adapts 4$\times$ faster than prior schemes. Being network-agnostic, it plugs into \emph{any} deep unfolding architecture, enabling field-deployable spectral imaging on drones, point-of-care scanners, and environmental sensors.

\bibliographystyle{IEEEtran}
\bibliography{ref}

\begin{figure*}[t]
    \centering
    \includegraphics[width=0.98\linewidth]{Fig_supp/harvard_supp_s2_1.pdf}
    \includegraphics[width=0.98\linewidth]{Fig_supp/harvard_supp_s2_2.pdf}
    \caption{Simulated HSI reconstruction comparisons on \texttt{Harvard} dataset. The left shows the scene measurement and the spectral curves corresponding to the selected region. Please zoom in for better view.
    }
    \label{fig:harvard1}
\end{figure*}

\begin{figure*}[t]
    \centering
    \includegraphics[width=0.98\linewidth]{Fig_supp/harvard_supp_s9_1.pdf}
    \includegraphics[width=0.98\linewidth]{Fig_supp/harvard_supp_s9_2.pdf}
    \caption{Simulated HSI reconstruction comparisons on \texttt{Harvard} dataset. The left shows the scene measurement and the spectral curves corresponding to the selected region. Please zoom in for better view.
    }
    \label{fig:harvard2}
\end{figure*}

\begin{figure*}[ht]
    \centering
    \includegraphics[width=0.98\linewidth]{Fig_supp/icvl_supp_s6_1.pdf}
    \includegraphics[width=0.98\linewidth]{Fig_supp/icvl_supp_s6_2.pdf}
    \caption{Simulated HSI reconstruction comparisons on \texttt{ICVL} dataset. The left shows the scene measurement and the spectral curves corresponding to the selected region. Please zoom in for better view.
    }
    \label{fig:icvl1}
\end{figure*}

\begin{figure*}[ht]
    \centering
    \includegraphics[width=0.98\linewidth]{Fig_supp/icvl_supp_s8_1.pdf}
    \includegraphics[width=0.98\linewidth]{Fig_supp/icvl_supp_s8_2.pdf}
    \caption{Simulated HSI reconstruction comparisons on \texttt{ICVL} dataset. The left shows the scene measurement and the spectral curves corresponding to the selected region. Please zoom in for better view.
    }
    \label{fig:icvl2}
\end{figure*}

\begin{figure*}[t]
    \centering
    \includegraphics[width=0.98\linewidth]{Fig_supp/real_supp_1.pdf}
    \includegraphics[width=0.98\linewidth]{Fig_supp/real_supp_2.pdf}
    \caption{\small Real-scene HSI reconstruction results.
    Shown are reconstructions of two real-world scenes across 4 selected spectral channels (out of 28), using various state-of-the-art methods. Compared to others, SlowFast-SCI restores finer details and better suppresses noise, highlighting its effectiveness in real-data scenarios.}
    \label{fig:app_real}
\end{figure*}

\end{document}